\documentclass[final,3p,times]{elsarticle}

\usepackage{amssymb}
\usepackage{graphicx}
\usepackage{pdflscape}
\usepackage{booktabs}
\usepackage{longtable}
\usepackage{algorithm}
\usepackage{algpseudocode}
\usepackage{hyperref}
\usepackage{float}
\usepackage{subfig}
\usepackage{amsmath}
\usepackage{multirow}
\usepackage{amsthm}

\journal{TBD} 

\begin{document}

\begin{frontmatter}



\title{From Overfitting to Robustness: Quantity, Quality, and Variety Oriented Negative Sample Selection in Graph Contrastive Learning} 

\author[inst1]{Adnan Ali}
\author[inst1]{Jinlong Li}
\author[inst1]{Huanhuan Chen}
\affiliation[inst1]{organization={University of Science and Technology of China},
            addressline={Jinzhai road}, 
            city={Hefei},
            postcode={23006}, 
            state={Anhui},
            country={China}}
\author[inst2]{Ali Kashif Bashir} 

\affiliation[inst2]{organization={Manchester Metropolitan University},
            addressline={M15 6BH}, 
            city={Manchester},
            country={United Kingdom}}

\begin{abstract}
Graph contrastive learning (GCL) aims to contrast \textit{positive-negative} counterparts to learn the node embeddings, whereas graph data augmentation methods are employed to generate these positive-negative samples. 
The variation, quantity, and quality of negative samples compared to positive samples play crucial roles in learning meaningful embeddings for node classification downstream tasks. 
Less variation, excessive quantity, and low-quality negative samples cause the model to be overfitted for particular nodes, resulting in less robust models. 
To solve the overfitting problem in the GCL paradigm, this study proposes a novel Cumulative Sample Selection (CSS) algorithm by comprehensively considering the quality, variations, and quantity of negative samples. 
Initially, three negative sample pools are constructed: easy, medium, and hard negative samples, which contain 25\%, 50\%, and 25\% of the total available negative samples, respectively. 
Then, 10\% negative samples are selected from each of these three negative sample pools for training the model. 
After that, a \textit{decision agent} module evaluates model training results and decides whether to explore more negative samples from three negative sample pools by increasing the ratio or keep exploiting the current sampling ratio. 
The proposed algorithm is integrated into a proposed graph contrastive learning framework named NegAmplify. NegAmplify is compared with the SOTA methods on nine graph node classification datasets, with seven achieving better node classification accuracy with up to  2.86\% improvement.
NegAmplify is available at: \url{https://github.com//mhadnanali/NegAmplify}.
\end{abstract}



\begin{keyword}
self-supervised learning \sep graph representation learning \sep graph contrastive learning \sep deep learning \sep negative sampling  
\PACS 0000 \sep 1111
\MSC 0000 \sep 1111
\end{keyword}

\end{frontmatter}


\section{Introduction}
\label{sec:Introduction}
Graphs are complex non-euclidean data structures, and learning meaningful graph and their element representations is a crucial foundation for applying machine learning to them, such as graph and node classification \cite{ZHANG2024111558}. 
Previous studies addressed this challenge by employing an iterative approach that propagates neighbor information until a stable fixed point is reached\cite{DBLP:journals/corr/abs-2102-10757}. However, this process was unreliable and computationally expensive \cite{BookHamilton}. 
Recently, Graph Representation Learning (GRL) gained the attention of data mining and machine learning communities as a way to exploit the richness of graph-structured data where graph neural networks are the main contributors \cite{MA2023110388}. 
GRL aims to embed nodes into low-dimensional vector space to preserve the features of nodes, neighboring information, and the graph structure \cite{GraphSageHamiltonYL17}.
It is performed in a supervised manner where various GRL methods are proposed in recent times \cite{GraphSageHamiltonYL17,KipfW16,velickovic2018graph}. 
However, a bottleneck of these supervised approaches is data annotation. 
The complex nature of graphs, massive data, expert knowledge, and manual dependency on humans make data annotation costly, impractical, slower, and not available all the time \cite{WANG2024111512}.
\par
Self-supervised learning (SSL) aims to solve the annotation crisis and learn the representation without labels \cite{graphsofdoi.13964,WANG2024111512}.  
The GRL domain has three types of SSL paradigms: Generative, Predictive, and Contrastive \cite{DBLP:journals/corr/abs-2006-10141}. 
Generative learning exploits the attributes and structure of graph data for self-supervision signal and aims to regenerate the graph or missing parts of graphs, nodes, or edges \cite{HGMAE2023}. 
Predictive methods try to predict the labels of unlabeled nodes using expert knowledge or statistical analysis. 
Lastly, contrastive learning is a data-data-based learning paradigm where data-data pairs are used for self-supervision signals, and the model learns by contrasting the difference between positive and negative samples \cite{LIANG2023156}. 
This study concentrates on contrastive learning in the graph domain, named Graph Contrastive Learning (GCL).
\par
GCL models are trained by contrasting positive and negative pairs, where positive pairs are similar or related data points, and negative pairs are dissimilar or unrelated\cite{10025823}.
These positive and negative samples are generated with data augmentation, and negative samples are uniformly selected to contrast with positive samples \cite{GCADBLP-abs-2010-14945}. 
Data augmentation is a well-studied area in the GCL domain, and there are various methods for data augmentation to create negative samples, including stochastic \cite{thakoor2021bootstrapped,DeepGrace2020,DAENS}, adaptive \cite{GCADBLP-abs-2010-14945,pmlr-v198-zhao22a,comunityGuo2023,10095350,FebAA.2207.01792}, and augmentation free \cite{AFGRL,AFGCL}. 
However, data augmentation is not the only critical part; the quality of generated and selected negative samples is vital for the GCL model's performance and needs attention \cite{cucoijcai}.
Negative sample selection was undermined before and is getting attention recently \cite{MIAO2022667,Yang_2023,10181235,hal-03575619}. 
The uniform selection performed in earlier methods \cite{DeepGrace2020,GCADBLP-abs-2010-14945} worked well for starters but offered two undermined gaps. 
\textit{First}, the uniform selection treats the same class nodes as negative samples (false negatives), and \textit{Second}, it bears a high negative to positive sample ratio (too many negative samples against few positive samples).   

\par
The \textit{first} gap is aimed in recent studies \cite{10181235,10382709,Zhang2024} and proposed to resolve by detecting the false negative samples from hard negative samples (samples which are difficult to distinguish from a positive sample). 
However, contrastive models are trained without label information; in such cases, identifying false negatives is effortful as no ground truth or annotation is available during training \cite{Zhang2024}.
The similarity metrics do not guarantee that identified false negative samples are false negatives rather than regular hard negative ones. 
We argue that these recent works are putting all effort into identifying the false negative samples\cite{10382709} without solid ground truth, and this process is an extra layer of computation with ambiguous results. 
Another approach to avoid false negatives is to generate negative samples, and some recent studies focus on generating counterfactual hard negative samples \cite{Yang_2023} or good negative samples \cite{MIAO2022667,hal-03575619}.
However, these techniques require excessive computation costs to generate negative samples or focus only on hard negative samples\cite{10382709}.
We argue that these approaches have given the negative sample selection problem a whole new dimension and currently only focus on hard negatives or false negative samples.
What if false negatives are not a problem? These \cite{Zhang2024} methods offer trivial performance improvements over uniform negative sample selection methods. 
A GCL model learns from variety, not from a specific type of samples (hard negatives), and that was the major reason for the success of models like GRACE \cite{DeepGrace2020} and GCA \cite{GCADBLP-abs-2010-14945} as they offered the uncontrolled variety by offering all nodes as negative samples. 

\par
The \textit{second} gap offered by uniform negative sample selection is the number of negative samples (quantity) as small (Cora) to large (Amazon Computers) datasets have a huge difference in the number of nodes and edge density, in such cases treating all nodes as negative samples against few positive pairs cause the overfitting, especially for large datasets. 
Overfitting is a challenge for both random negative sample selection approaches and hard negative sampling \cite{10382709}. 
Training the model with the same type of negative samples (too many hard negatives) does not provide generalization \cite{ABGML}, or training with all nodes as negative samples at every epoch increases the chances of overfitting \cite{Yang_2023}. 
Furthermore, if the model updates the embeddings with backpropagation at every epoch, why use all negative samples simultaneously? 
CuCo\cite{cucoijcai} solves this problem by training the model with Curriculum Contrastive Learning. However, CuCo is a graph-level framework, and curriculum learning implications are unknown in the node classification domain. 

\par
We propose to cover these gaps and build our motivation on three elements. 
1- \textit{Quantity}: Offering all nodes as negative samples causes overfitting, as the imbalance can lead the model to prioritize distinguishing between a large number of negative samples rather than capturing the essential characteristics of the positive sample. 
2- \textit{Quality}: False negatives are not as big an issue as stated in previous work \cite{10382709,10181235,Zhang2024}, the model still can learn from false negatives as the model also required to distinguish between nodes from the same class for unseen data. 
A model trained with only true positives will have a biased representation and limited robustness for the unseen data. 
3- \textit{Variety}: Considering only a particular type of negative sample undermines the variety. The model learns to contrast specific node types and loses the robustness for unseen data.  



\par
To consider three elements: quality, quantity, and variety of negative samples, we propose a graph contrastive learning framework ``\textit{NegAmplify}: Conditional Increase in Negative Samples".
NegAmplify controls the quality by dividing the negative samples into easy, medium, and hard negative sample pools. 
It contains a negative sample selection algorithm ``Cumulative Sample Selection (\textit{CSS})", to control the variety and quantity based on negative sample pools. 
Below are the contributions of this study:

\begin{enumerate}	 
\item We introduce a negative sample selection algorithm named \textit{Cumulative Sample Selection (CSS)} to control the quantity of negative samples to the model while offering the variety at each epoch. Based on previous epochs' contrastive losses, a \textit{decision agent} from CSS decides whether the number of negative samples should be increased or keep exploiting the current quantity while offering different negative samples to the model. 

\item We divide the negative samples into three pools: easy, medium, and hard negative sample pools, where negative samples belonging to the easy pool are easy to distinguish from positive samples, medium and hard have medium and hard level complexity to distinguish from positive samples. 
This pool divining offers the flexibility to control the quality and variety of negative samples. CSS is offered negative samples from these pools. 
 
\item We propose a graph contrastive learning framework \textit{NegAmplify}: Conditional Increase in Negative Samples, which deploys CSS and negative sample pools in a graph contrastive learning paradigm. NegAmplify improves the node classification accuracy of seven of nine benchmark datasets with up to 2.86\%. 

\item We empirically analyze the percentage of negative samples to get the best results against benchmark datasets. Results conclude that dense datasets require fewer negative samples than sparse datasets. 

\item We compare the results of NegAmplify CSS with the curriculum learning paradigm and random selection of negative samples rather than from easy, medium, and hard negative sample pools. Curriculum learning and random selection still perform better than the SOTA method on six datasets, but the NegAmplify CSS version takes the lead overall. 
\end{enumerate}

\section{Literature Review}
\label{sec:c3RelatedWork}
Negative sampling was first introduced in the context of word embeddings in Word2Vec\cite{word2vec} and has since been adapted to various domains, including computer vision \cite{DBLP:journals/corr/abs-2002-05709} and graph machine learning \cite{ContrastiveGenerative}.
DeepWalk \cite{Deepwalk32} employs a methodology similar to the Skip-gram model used in Word2Vec, which includes the concept of negative sampling for efficient training, making it one of the earliest papers to use negative samples in graph machine learning. 
Later, Graph Convolutional Networks (GCNs) \cite{KipfW16} laid the groundwork for contrastive learning methods in graph neural networks, and now negative samples are used in graph contrastive learning \cite{negativeReview}.
Negative sampling is a crucial element of graph contrastive learning that impacts the quality of the learned representations \cite{HardNegativerobinson2021}. 
By contrasting positive and negative samples, the model learns to distinguish between different data patterns \cite{Zhang2024}. 
Training the model without negative samples (with only positive samples) leads to overfitting, low-level embeddings, and trivial performance improvements \cite{10382709}. 

Negative samples are significant in contrastive learning, but previous work has utilized different methodologies to determine the variety, quantity, and quality of negative samples \cite{ContrastiveGenerative}. 
Such selections are either performed in \textit{Random} manner or with some approaches like \textit{Hard Negative Mining}, a combination of \textit{Hard and Easy}, and \textit{Adversarial} manner. 
This section critically reviews the different approaches used in the literature to select the negative samples, discusses the differences in these methods, and reviews their usage in respective studies.

\subsection{Random Sampling}
The most direct approach in negative sample selection is a random or uniform selection of negative sampling, where samples are chosen indiscriminately from the graph, without following any specific criteria \citep{Deepwalk32, node2vec}. 
It is a straightforward approach requiring less processing time, and it is independent of auxiliary data, which can be beneficial in scenarios with limited information about downstream tasks and datasets, which makes it easy to generalize on various downstream tasks using different datasets \cite{xia2022progcl}. 
In practice, random sampling is frequently utilized as a default strategy in computer vision \cite{DBLP:journals/corr/abs-2002-05709} and graph representation learning \cite{GCADBLP-abs-2010-14945}. 
However,  the downside of this strategy lies in its potential to produce low-quality negative samples and excessive memory usage, as the random selection does not guarantee the representativeness or relevance of the samples to the specific task or model \cite{zhao2021graph}.
\par
DeepWalk \cite{Deepwalk32} uses random negative sampling for learning social network representation. 
Another approach falling inside the domain is the selection of all possible nodes as negative samples, and utilized by GRACE  \cite{DeepGrace2020}, GCA \cite{GCADBLP-abs-2010-14945}, COSTA \cite{CostaZhang_2022} and in PyGCL empirical study \cite{zhu2021empirical}. 
Random selection might not always be effective, as some negative samples can be more informative than others \cite{10025823}. 
However, the approach does not provide a method for deciding which negative samples are better for the model. 
The biggest drawback of this approach is false negative samples, which downgrade the performance, and removing such false negative samples can increase the model efficiency \cite{10181235,10382709,Zhang2024}.

\subsection{Hard Negative Mining}
Hard negative are close to positive samples in the embedding space but belong to different classes or contexts \citep{Schroff_2015_CVPR}. 
The uniform negative sampling strategy limits the expressive power of contrastive models, as not all negative nodes provide meaningful knowledge for effective representation learning \cite{10181235}.
In contrast to random sampling, hard negative mining has emerged as a vital aspect, particularly for enhancing the effectiveness of contrastive learning models \cite{HardNegativerobinson2021}. 
This approach involves selecting or generating negative samples that are challenging to discriminate from positive samples, aiming to improve the learning robustness and performance \cite{s23125572}.
However, without class information, finding the negative samples is challenging, and it leads to the inclusion of false negatives \cite{xia2022progcl}. 
\par

\par
GraphCL\cite{hal-03575619} investigates three strategies (random, feature-based and graph-based) to select the best among them, where feature-based and graph-based also aim to find the hard negative samples. 
HSAN \cite{liu2023hard} addresses two critical issues in hard sample mining: the neglect of structural information in hardness measurement and the focus solely on hard negative samples, ignoring hard positives. 
It incorporates a comprehensive similarity measure and a dynamic sample-weighing strategy. This approach effectively mines both hard negative and positive samples, improving the discriminative capability of the network. However, it is a clustering paper, not a classification.  
ProGCL \cite{xia2022progcl} proposes a hard negative sampling mechanism that evaluates the potential informativeness of negatives. 
It avoids the pitfalls of selecting overly challenging negatives, which could lead to noise overfitting, or overly simple negatives, which offer little to no learning signal. 
GC4SRec \cite{s23125572} introduces a novel approach that leverages graph neural networks for deriving user embeddings. This methodology, critical in the field, encompasses an encoder designed to calculate the importance scores of items. 
Furthermore, the study innovatively applies several data augmentation techniques to create a contrasting view grounded in these importance scores. 
\par
The study \cite{MIAO2022667} that uses a supervised signal to identify the false negatives; although it improved the node classification results, such approaches are not self-supervised. 
The AdaS \cite{10181235} framework is designed to adaptively encode the importance of negative nodes, fostering learning from the most informative nodes in the graph. Additionally, it introduces an auxiliary polarization regularizer to mitigate the negative effects of false negatives, thereby boosting AdaS's discriminative capabilities.
AUGCL \cite{10382709} focuses on improving hard negative mining through an uncertainty learning framework and   uniquely emphasizes uncertainty estimation to assess the hardness of negative instances, offering a fresh perspective in contrastive learning within the graph domain.
Another study \cite{Yang_2023} generates high-quality negative samples for graph contrastive learning using a counterfactual mechanism. This approach aims to produce negative samples similar to positive ones but with different labels, ensuring the negatives are both hard and true. This is also a graph classification method.
Methods like these can face challenges like computational complexity, potential overfitting to hard negatives, or difficulties in accurately assessing the uncertainty or hardness of negatives in complex graph structures. 
\par
Hard negative mining also causes the selection of the same class nodes as negative samples (false negatives).
A hurdle to finding false negatives in graph contrastive learning is that label information can not be used, so it is impossible to know if nodes are from the same class or are different. 
Hard negative sampling in graph contrastive learning is not a very mature area; it is still in its infancy.
Furthermore, the majority of the research is focused on finding the false negatives. 

\subsection{Hard and Easy}
While hard negative samples are pivotal in training models, completely disregarding easy negative samples is not advisable \cite{cucoijcai}.
Easy negatives, as demonstrated by the study\cite{8578392}, can be instrumental in producing hard negatives and also play a crucial role in maintaining stability during the initial stages of training.
As the training progresses, incorporating a blend of both easy and hard negatives, at various levels of difficulty, can facilitate more nuanced representations. 
HDML\cite{DBLP:journals/corr/abs-1903-05503} dynamically adjusts the sampling of negative pairs based on their relative hardness. This method effectively balances the use of hard and easy negatives, improving the discriminative power of the model.
\par
The concept of curriculum learning in negative sampling, exemplified by CuCo \cite{cucoijcai}, follows a progressive approach by sorting candidates based on a scoring function and gradually moving from easier to more challenging samples.
This method demonstrates that varying the difficulty of negative samples can benefit model optimization across different domains. 
GNNO \cite{Fan_2023} is a graph classification negative sampling approach based on Neighborhood Overlap, utilizing curriculum learning to control negative sample hardness, consistently outperforming existing strategies in sequential recommendation.
However, the first two methods \cite{8578392,DBLP:journals/corr/abs-1903-05503} mentioned are from computer vision, while CuCo \cite{cucoijcai} and GNNO \cite{Fan_2023} are graph classification papers. 
We did not find any paper that uses both hard and easy negative samples in graph contrastive learning node classification. 
Although using both hard and easy negative samples could be effective if they are used properly for node classification, a research gap has not yet been considered.

\subsection{Adversarial Negative Sampling}
These approaches are inspired by GANs (generative adversarial networks) and generate negatives that are specifically designed to be difficult for the model to differentiate from positives \cite{DBLP:journals/corr/abs-1811-04155}.  
It is an approach to creating negative samples that helps the model learn useful embeddings, which can also be applied to create hard negative samples. 
In graph contrastive learning, adversarial negatives can be constructed by perturbing the graph structure or node features \citep{negativeReview}.
By exposing models to a broader range of difficult samples, adversarial negative sampling helps in achieving more robust and effective learning outcomes \cite{8578392}. 
However, just hard samples may not be enough as a model should have a variety from easy to hard, as real-world datasets may have different kinds of nodes or features. 
We argue that graph contrastive learning is already a computation-hungry task, and adversarial methods further increase the complexity with marginal improvements that can already be achieved with random samplings. 
One big drawback of the adversarial approaches is that there is open space for adversarial attacks where graph contrastive learning is already prone to such attacks \cite{feng2022adversarial}.
It provides a solution for negative sampling but opens another door for attacks.



\par

\par
In summary of the literature review, although there are multiple negative sample approaches from random, uniform selection hard negative sample-based approaches, then curriculum learning-based approaches. 
The available hard negative mining approaches are majorly focused on finding the false negatives. 
We argue that the same type of negative samples can cause overfitting as the model is trained for only hard tasks while the real-world data is not focused on a specific type. 
There is a research gap in designing a negative sample selection approach that maintains negative samples' quality, quantity, and variety while consuming less memory and computation resources. 

\section{Methodology}
\label{sec:c3:methdologyNGCL}
\begin{figure}[h]		
	\centering
	\includegraphics[height=0.5\textheight, width=0.3\textheight]{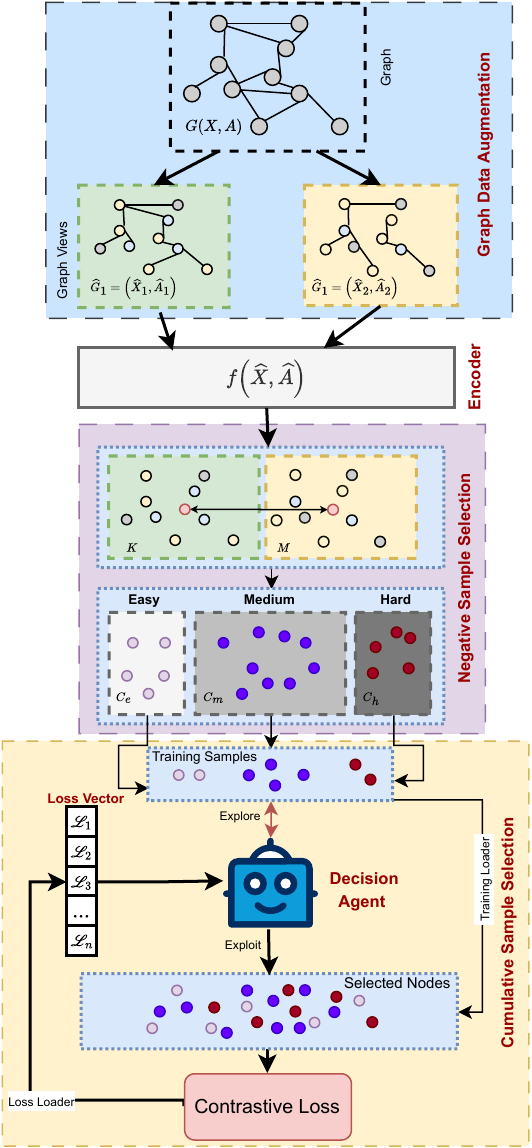}
	\caption{Model diagram of NegAmplify.}
	\label{fig:c3:NegModelDiagram}       
\end{figure}

This section introduces a model called: ``\textit{NegAmplify}: Conditional Increase in Negative Samples".  NegAmplify refers to a negative samples amplification method that increases negative samples based on certain algorithms.  
Figure \ref{fig:c3:NegModelDiagram} illustrates the model diagram of \textit{NegAmplify}. 
It has four major modules: 1- \textit{Graph Data Augmentation}, 2- \textit{Encoder}, 3- \textit{Negative Sample Selection}, and 4- \textit{Cumulative Sample Selection}. 
First module \textit{Graph Data Augmentation} creates the corrupted copies of the original graph. The second module \textit{Encoder} encodes the graph data to embeddings. 
Third module \textit{Negative Sample Selection} identifies the positive samples and groups the negative samples into pools of Easy, Medium, and Hard negative samples. At last, \textit{Cumulative Sample Selection} is a proposed algorithm to choose the representative negative samples from module three's negative sample pools and define criteria to increase the number of negative samples based on given conditions. 
A detailed description of each module is provided in the subsections below. 
	
\subsection{Graph Data Augmentation}
Data augmentation is the first module of \textit{NegAmplify}, which creates augmented graph views. 
Figure \ref{fig:c3:NegModelDiagram} illustrates that graph $G=(X, A)$ converted to two graph views $\hat{G}_1=(\hat{X}_1, \hat{A}_1)$ and $\hat{G}_2=(\hat{X}_2, \hat{A}_2)$. The applied augmentation methods are feature masking at the feature level and edge dropping at the topology level, where further elaboration is below: 
 
\subsubsection{Edge Removing}
We randomly remove the portion of existing edges from the graph to corrupt the graph at the structural level. 
First, we create a masking matrix $M_m\in\{0,1\}^{N \times N}$ where $N$ depicts the number of nodes and $M_m$ is comprised of binary elements. 
Entries of $M_m$ are drawn with Bernoulli distribution $B(1-E_r)$ where $E_r$ is the probability of removing edges and is a hyper-parameter.  The resulting masking matrix is element-wise multiplied with adjacency matrix $A$ as given in Equation \ref{eq:c3:edgeReoming}:
\begin{equation}
        \label{eq:c3:edgeReoming}
	\hat{A}=A \circ M_m
\end{equation}
where `$\circ$' is element-wise multiplication or Hadamard product. 
	
\par
We only remove edges and do not add new edges as replacements for removed edges. As a result, the augmented graph should become sparser than before or may have disconnected subgraphs/nodes.
Figure \ref{fig:c3:NegModelDiagram} illustrates the edge removing where augmented graph $\hat{G}_1$ and $\hat{G}_2$ has fewer edges than $G$. It should also be noted that $\hat{G}_2$ is divided into two subgraphs, depicting that edge removing can cause disconnection between nodes, and some nodes can lose their neighborhood.

\subsubsection{Feature Masking}
Feature masking refers to masking elements of the feature matrix with zeros. 
Formally, a random vector $\nu \in \{0,1\}^F$ is created where $F$ depicts the number of features in the feature matrix. $\nu$ is a binary vector where entries are drawn with Bernoulli distribution $B(1-F_m)$ where $F_m$ is the probability to mask features and control how many entries will be 0 or 1. It is also a hyper-parameter setting. 
Then, each entry of $\nu$ is multiplied with columns of the feature matrix, where 1 makes no changes and 0 masks the whole column of the feature matrix. In previous work \cite{DAENS} defines such masking as 1D masking and it is presented in Equation \ref{eq:c3:fmasking}:
\begin{equation}
	\label{eq:c3:fmasking}
	\hat{X}=X  \circledcirc \nu
\end{equation}
where $ \circledcirc$ presents that each element of $\nu$ will be multiplied with corresponding column of feature matrix $X$. 
Feature masking is illustrated in Figure \ref{fig:c3:NegModelDiagram}, where changed colors of nodes in $\hat{G}_1$ and $\hat{G}_2$ depicts that nodes are not the same as before and their some features have been masked. 
These (Edge dropping and Feature masking) methods combined create the augmented graph given that in Equation \ref{eq:c3:dataAug}: 
\begin{equation}
	\label{eq:c3:dataAug}
	\hat{G}= (\hat{X}, \hat{A})
\end{equation}
We apply stochastic augmentation functions following the SOTA methods \cite{DeepGrace2020,thakoor2021bootstrapped,velikovi2019deep}  to elaborate the contribution of negative sampling rather than focusing on graph data augmentation.

\subsection{Encoder}
We employ  two layers of GCN \cite{KipfW16} as the encoder. 
Figure \ref{fig:c3:NegModelDiagram} illustrates the role of the encoder in between augmented views and node embeddings, where the role of the encoder is to learn the node representations in the embeddings space.
\par  
Formally, let \( X \) be the feature matrix where each row corresponds to a node feature vector. 
The GCN Encoder comprises layers that update node representations by aggregating information from their neighbors. 
For the \( i \)-th view, the layer-wise update at the \( l \)-th layer is given by:

\begin{equation}
	H^{(l+1)(i)} = \sigma\left(\tilde{D}^{(i)-\frac{1}{2}} \tilde{A}^{(i)} \tilde{D}^{(i)-\frac{1}{2}} H^{(l)(i)} W^{(l)}\right)
\end{equation}

where:
\begin{itemize}
	\item \( \tilde{A}^{(i)} = A^{(i)} + I_N \) is the adjacency matrix of the \( i \)-th view with added self-loops.
	\item \( \tilde{D}^{(i)} \) is the degree matrix of \( \tilde{A}^{(i)} \).
	\item \( H^{(l)(i)} \) represents the hidden features from the \( i \)-th view at the \( l \)-th layer, with \( H^{(0)(m)} = X \) being the input features.
	\item \( W^{(l)} \) is the weight matrix for the \( l \)-th layer shared across views.
	\item \( \sigma \) is a non-linear activation function, such as ReLU.
\end{itemize}

The cross-view aggregation at the \( l \)-th layer can be represented as:

\begin{equation}
	H^{(l+1)} = \text{AGGREGATE}\left(\{H^{(l+1)(i)}\}_{i=1}^I\right)
\end{equation}

where AGGREGATE is an aggregation function that combines the representations from all views.

\subsection{Negative Sample Selection}
This subsection aims to choose the representative subset of negative samples from $Neg$ for training rather than selecting all nodes, where $Neg$ is a set of all possible negative samples. 
\subsubsection{Learned Representation Nodes Pool}
The encoder converts the nodes into representations, and nodes are no longer graph nodes but are converted into embeddings or node representations. 
Figure \ref{fig:c3:NegModelDiagram} presents the node embeddings from each graph view as $K$ and $M$. 
The $K$ and $M$ are representations sets  of graph views $\hat{G}_1$ and $\hat{G}_2$ learned based on contrastive loss. 
Below, we define positive and negative samples. 	
\paragraph{Positive Samples}
Let positive samples $Pos=(k_i,m_i)$ a pair of nodes which is a copy of the same nodes in two different representations ($K$, $M$) of views ($\hat{G}_1$, $\hat{G}_2$ ). 
Figure \ref{fig:c3:NegModelDiagram} presents the positive sample nodes with pink color nodes and points them with arrows to distinct from negative samples. 

\paragraph{Candidate Negative Samples}
All of the nodes from both views are candidates for negative samples and are presented as $Neg$ except positive samples. 
Candidate negative samples are presented in the equation: 
\begin{equation}
	\label{eq:c3:negsamples}
	Neg= (K \cup M) - Pos 	
\end{equation}
where $K$ and $M$ are node representations and $Pos$ are positive samples. 
$Neg=(ki,\{m_1,m_2,\ldots,m_i\})_{m_i \neq k_i}$
These nodes $Neg$ qualify to be negative samples, but we do not consider all of them for training.	
	
\subsubsection{Nodes Hardness Pools}	
This section divides candidate negative samples into three subsets based on their hardness to positive samples. 
The subsets are Easy, Medium, and Hard negative samples, where cosine similarity is the hardness measuring metric. We choose cosine similarity for comparison based on CuCo \cite{cucoijcai}, ProGCL \cite{xia2022progcl}, and InfoNCE loss. 
\par
Formally, let $Neg$ be a set of candidate negative nodes (Equation \ref{eq:c3:negsamples}), where each node at this level is an embedded vector value comprised of features of nodes and their neighbors (via aggregation).  We calculate the similarity between each node with Equation \ref{eq:c3:cosinesim}: 
	\begin{equation}
		\label{eq:c3:cosinesim}
		\text{Cosine Similarity}(Pos, Neg) = \frac{\sum_{x \in Pos} \sum_{y \in Neg} x \cdot y}{\sqrt{\sum_{x \in Pos} x^2} \cdot \sqrt{\sum_{y \in Neg} y^2}}
	\end{equation}
where \textit{Cosine Similarity}$(Pos, Neg)$ is the cosine similarity between the positive and negative pairs. 
However, there are multiple negative samples. So we calculate the $\text{Cosine Similarity}(Pos, Neg)$ and add it into cosine similarity vector $c$ as per Equation \ref{eq:c3:cosinesimvec}.  
\begin{equation}
	\label{eq:c3:cosinesimvec}
	C=c\parallel\text{Cosine Similarity}(Pos, Neg)
\end{equation}
where $\parallel$ presents the appending. Now, $C$ has the cosine similarity of each negative node with positive samples.
\par 
The next step is sorting the candidate negative samples per their cosine similarity. 
We apply a sorting operation to sort the cosine similarity in descending order, where lower similarity depicts that the negative sample is different from the positive sample, and high similarity depicts that both nodes are highly similar and hard to distinguish. 
Equation \ref{eq:c3:sorting} presents the sorting method. 
	
\begin{equation}
	\label{eq:c3:sorting}
	C = \text{sorted}(C)
\end{equation}
The output of Equation \ref{eq:c3:sorting} is a sorted list of candidate negative samples presented as $C$. 
The vector $C$ is used in the next subsections to create a set of Easy, Medium, and Hard negative samples. 

\textbf{Easy Negative Samples:}
The negative samples are easy to distinguish from positive samples. 
A subset of vector $C$ presented as $C_e$, where $C_e \subset C$. 
Let $C$ be a set of sorted candidate negative samples where $C=\{c_1,c_2,c_3,\ldots,c_n\}$, where $C_1$ has the lowest similarity with the positive sample and  $C_1$ has a second lowest and so on. 
\par	
We define, $C_e$ as first 25\% elements of $C$ in the Equation \ref{eq:c3:easy}:
	\begin{equation}
		\label{eq:c3:easy}
		C_e = \{c_i \mid i \leq \lceil 0.25 \cdot n \rceil \}
	\end{equation}
where $c_i$ is the $i$-th element of set C, $n$ represents the total number of elements in $C$ and $\lceil 0.25 \cdot n \rceil$ calculates the ceiling value of $ 0.25 \cdot n$, giving the first 25\% elements of $C$.

\textbf{Medium Negative Samples:}
Medium Negative Samples are those that are medium in similarity to positive samples. 
They are not very hard or easy to distinguish from positive samples. 
These are the middle 50\% elements from $C$ and presented as $C_m$, while $C_m \subset C$. 
Equation \ref{eq:c3:medium} presents the medium negative samples: 
\begin{equation}
	\label{eq:c3:medium}
	C_m = \{c_i \mid i > \lceil 0.25 \cdot n \rceil \text{ and } i \leq \lceil 0.75 \cdot n \rceil \}
\end{equation}
where $c_i$ represent the $i$-th elements of set $C$, $n$ represents the total elements in $C$, $ i> \lceil 0.25 \cdot n \rceil$ excludes the starting 25\% elements, and $i \leq \lceil 0.75 \cdot n \rceil$ excludes the last 25\% elements. 
	
\textbf{Hard Negative Samples:}
Hard Negative Samples are samples that are hard to distinguish from positive samples. 
These are the opposite of easy negative samples and are the last 25\% elements of $C$, presented as $C_h$ where $C_m \subset C$.
Equation \ref{eq:c3:hard} presents the hard negative samples. 
\begin{equation}
	\label{eq:c3:hard}
	C_h = \{c_i \mid i > \lceil 0.75 \cdot n \rceil\}
\end{equation}
where $c_i$ represent the $i$-th elements of set $C$, $n$ represents the total elements in $C$, and $i > \lceil 0.75 \cdot n \rceil$ are last 25\% elements.	
	
Overall, $C$ is a union of all the types of these methods presented as $C=C_e \cup C_m \cup C_h$, and there are no common elements between them.
Given that the intersection between these three sets is empty set $C_e \cap C_m \cap C_h= \{\}$. 
A combined formal definition for easy, medium, and hard negative samples is in Equation \ref{eq:c3:easyhardneg}:

\begin{equation}
		\label{eq:c3:easyhardneg}
		\{C_e, C_m, C_h\},  = \begin{cases}
			\{c_i \mid i \leq \lceil 0.25 \cdot n \rceil \}, & \text{for the first 25\% elements} \\
			\{c_i \mid i > \lceil 0.25 \cdot n \rceil \text{ and } i \leq \lceil 0.75 \cdot n \rceil \}, & \text{for the 26\% to 75\% elements} \\
			\{c_i \mid i > \lceil 0.75 \cdot n \rceil \}, & \text{for the last 25\% elements}
		\end{cases}
\end{equation}

We use a combination of $C_e, C_m, and C_h$ for training; however, the Cumulative Sample Selection is presented in the next section. 
	
\subsection{Cumulative Sample Selection}
This section proposes a Cumulative Sample Selection (CSS) method that takes $C_e, C_m, C_h$ as input and selects representative negative samples to train the model and generate embeddings useful for downstream tasks. 
Figure \ref{fig:c3:NegModelDiagram} presents the graphical illustration of the CSS, whereas  Algorithm \ref{Algo:NegAmplify} presents the working. 
\par
The subsections of CSS are: \textit{Training Samples} are used for training,  \textit{Contrastive Loss} is to calculate the loss, \textit{Loss Vector} to keep track of calculated loss,  \textit{Decision Agent} determines to add more negative samples or not, while the \textit{Selected Nodes} are positive and negative nodes given to contrastive loss to calculate the contrastive loss of current epoch. 
Below, we explain how these subsections work. 

\subsubsection{Training Samples}
Figure \ref{fig:c3:NegModelDiagram} illustrates that \textit{Training Samples} holds the nodes used for training. 
It takes negative samples from $C_e, C_m, C_h$ and passes them to \textit{Selected Nodes} pool via  Training Loader. 
The size of training samples is decided with a variable  $\kappa$, which has an initial value depicting the starting percentage to train the model.   
The $\kappa$ value is consistent for all easy $C_e$, medium $C_m$, and hard $C_h$ negative samples. 
Let $\kappa=10\%$ indicate that 10 percent of negative samples will be selected from easy $C_e$, medium $C_m$, and hard $C_h$ negative samples.

\subsubsection{Contrastive Loss}
The contrastive loss aims to ensure that positive samples (similar node pairs) are brought closer together in the representation space while negative samples (dissimilar node pairs) are pushed apart
Equation \ref{eq:c3loss} presents the contrastive loss used to train the NegAmplify. 


\begin{equation}
\label{eq:c3loss}
\mathcal{L} = -\mathbb{E}_{(k, m^+, \{k^-\}, \{m^-\})} \left[ \log \frac{e^{(k \circ m^+ / \tau)}}{e^{(k \circ m^+ / \tau)} + \sum_{k^-} e^{(k \circ k^- / \tau)} + \sum_{m^-} e^{(k \circ m^- / \tau)}} \right]
\end{equation}

where  $\mathcal{L}$ represents the contrastive loss, $\mathbb{E}$ is the expectation over the set of all positive and negative pairs; $(k, m^+, {m^-})$ denotes a node representation $k$ from the first graph view, its positive counterpart $m^+$ from the second graph view, and a set of negative samples $m^-$, $e^{}$ represents the exponential function. Furthermore, $\circ$ denotes the dot product; $\tau$ is the temperature parameter to control the sharpening or softening of the probability distribution. A higher value of $\tau$ makes the distribution softer, while a lower value makes it sharper. $\log$ is the natural logarithm; the summation $\sum_{k^-}$ sums over all intra-view negative samples and $\sum_{m^-}$ sums over all inter-view negative samples.
The denominator includes sums over both ${k^-}$ and ${m^-}$, while $\sum_{k^-} e^{(k \cdot k^- / \tau)}$ represents the sum of the exponentiated similarities between the node representation $k$ and each of the negative samples in the $K$ set (intra-view).
Finally, $\sum_{m^-} e^{(k \cdot m^- / \tau)}$ represents the sum of the exponentiated similarities between $k$ and each of the negative samples in the $M$ set (inter-view).
The loss function aims to maximize the similarity between positive pairs (similarity scores closer to 1) and minimize the similarity between negative pairs (similarity scores closer to 0). 

\subsubsection{Loss Vector}
Loss is calculated at each training epoch where loss vector $\vec{\mathcal{L}}$ is a queue-like vector that stores the last $n$ epochs loss and discards the old ones.
Loss vector $\vec{\mathcal{L}}$ is a vector that stores the contrastive loss $\mathcal{L}$ values. 
Figure \ref{fig:c3:NegModelDiagram} illustrates that \textit{Loss Loader} passes the value of the latest loss to Loss vector $\vec{\mathcal{L}}$.

\subsubsection{Decision Agent}
The decision agent is a module that decides whether to keep exploiting the current percentage of selected negative samples or explore and increase the percentage of negative samples. 
It takes the decision based on $\kappa_{max}$ and Loss vector $\vec{\mathcal{L}}$ as illustrated in Equation \ref{eq:c3:exmporeorexploit} the mathematical notation. 

\begin{equation}
    \label{eq:c3:exmporeorexploit}
    \breve{Ag} = \begin{cases}
        \text{Exploit } & \text{if } \left((\sum_{i=n+1}^{N}  \bar{l}_i + \xi) < \sum_{i=1}^{n} \bar{l}_i\right) \text{ and } \left(\kappa >\kappa_{\text{max}}\right) \\
        \text{Explore  } & \kappa=\kappa+1
    \end{cases}
\end{equation}
where $\sum_{i=n+1}^{N} \bar{l}_i$ is sum of last $n$ values in $\vec{\mathcal{L}}$.
$N$ is the length of $\vec{\mathcal{L}}$. $\xi$ is the constant of change that decides the loss is increasing over epochs.  $\sum_{i=1}^{n} \bar{l}_i$ presents the first $n$ values in $\vec{\mathcal{L}}$. $\kappa$ is the current percentage of negative samples used in training. $\kappa_{max}$ is the maximum percentage that can be used to train the model. $\breve{Ag}$ is agent which takes the decision to \textit{Exploit} or \textit{Explore}.
\textit{Exploit} indicates that keep using the same percentage of negative samples (referring to that loss is decreasing). In contrast, \textit{Explore} refers that loss is not increasing anymore, time to increase the percentage of negative samples. 

\subsubsection{Selected Nodes}
Figure \ref{fig:c3:NegModelDiagram} illustrates the selected nodes as per the decision of Agent $\breve{Ag}$. 
These nodes are used to train the model and calculate the contrastive loss. 
Algorithm \ref{Algo:NegAmplify} presents the working of Cumulative Sample Selection (CSS).  
	
\begin{algorithm}[t]
\begin{algorithmic}[1]
\caption{Working of Cumulative Sample Selection (CSS)}
\label{Algo:NegAmplify}
\State $Pos=\{\text{Positive Sample}\}$
\State $C_e \gets \{ c_e^1 , c_e^2,c_e^3,\ldots,c_e^n\}$  \Comment{Easy Negative samples}
\State $C_m \gets \{ c_m^1 , c_m^2,c_m^3,\ldots,c_m^n\}$ \Comment{Medium Negative samples}
\State $C_h \gets \{ c_h^1 , c_h^2,c_h^3,\ldots,c_h^n\}$
\Comment{Hard Negative samples}
\State $\kappa \gets 10\%$

\For {$epochs \geq n$}:
\State $\Ddot{N}_e \gets \text{sample}(C_e, initial\times |C_e|)$
\State $\Ddot{N}_m \gets \text{sample}(C_m, initial\times |C_m|)$
\State $\Ddot{N}_h \gets \text{sample}(C_h, initial\times |C_h|)$
\State $\Ddot{N} = \Ddot{N}_e \cup \Ddot{N}_m \cup \Ddot{N}_h $   \Comment{Selected Nodes}
			
\State $\mathcal{L}$=Loss($Pos$,$\Ddot{N}$) \Comment{Contrastive Loss}
\State $\vec{\mathcal{L}}$ = $\vec{\mathcal{L}}$.append($\mathcal{L}$)

\If{$(\sum_{i=n+1}^{N}  \bar{l}_i + \xi) < \sum_{i=1}^{n} \bar{l}_i \text{ and } \kappa >\kappa_{max}$}
\State Exploit \Comment{Keep using the same percentage of Negative Samples}
\ElsIf{$\kappa$<=$\kappa_{max}$}
\State $\kappa$= $\kappa$+1\%  \Comment{Explore new negative samples}
\EndIf
			
\EndFor
			
\end{algorithmic}
\end{algorithm}

\section{Experimental Studies and SOTA Comparison}
\label{sec:c3Results}
This section compares our proposed method \textit{NegAmplify} with state-of-the-art self-supervised learning methods.   
Our experimental framework is carefully designed to evaluate the performance of our proposed model NegAmplify rigorously.
The subsections are as follows: 
The \textit{Evaluation Setup} used for experiments, \textit{Datasets} presents the benchmark datasets used for evaluation, \textit{Baselines} are SOTA method used for comparison, \textit{Implementation Details} present the software, hardware, hyperparameter settings to recreate the results, \textit{NegAmplify Comparison with SOTA} presents the comparison analysis wit SOTA methods.
Finally,  \textit{Percentage Parameter Analysis} analyzes the maximum percentage of negative samples to get the best results.

\subsection{Evaluation Setup}
Our evaluation setup follows the SOTA methods \cite{thakoor2021bootstrapped,GCADBLP-abs-2010-14945,velikovi2019deep}, and we adopted the linear evaluation scheme for all experiments. 
The classifier's performance is assessed using the micro-averaged F1-score, following a transductive task setting with a data split of 10\% for training, 10\% for validation, and 80\% for testing.
The models are trained in an unsupervised manner without any label usage during the training phase. 
The resulting output embeddings are then utilized for training and testing a logistic regression classifier with $l_2$ regularization. 
Details about the implementation and hyper-parameter configurations can be found in \ref{sec:c3:implementation} of this study for replicability of the results.

\subsection{Dataset}
We employ diverse benchmark graph datasets for extensive experiments and Table \ref{tab:c3:DatasetDetails} presents the stats of these datasets. 
The dataset can be divided into two parts based on their sparsity. 
Sparse datasets are  Planetoid \cite{CiteseerDataset, CoraDataset, PubMedDataset} and DBLP\cite{dblpdataset}, while the remaining datasets are dense datasets. 
These two groups have similar properties at the sparsity level; however, the features are very different. 
For example, CiteSeer and PubMed are both sparse datasets; however, the number of nodes, features, and classes differs greatly.
Similarly, Coauthor CS and Amazon photos are dense datasets; however, the node degree, number of features, and classes differ greatly. 
The combination of these nine datasets covers all possible features, like binary or non-binary features, isolated nodes, number of classes, features, edges, etc, which vary in these datasets.

\begin{table}
	\centering
	\caption{Properties of classification benchmark datasets used in this work. }
	\label{tab:c3:DatasetDetails}       
	\begin{tabular}{llllll}
		\hline\noalign{\smallskip}
		Parent&Dataset & Nodes & Features &Edges& Classes  \\
		\noalign{\smallskip}\hline\noalign{\smallskip}
		Planetoid&Cora&2708&1433&10556&7\\
		Planetoid&CiteSeer&3327&3703&9104&6\\
		Planetoid&Pubmed&19717&500&88648&3\\
		DBLP&DBLP&17716&1639&105734&4\\
		WikiCS&WikiCS&11701&300&216,123&10\\
		Amazon&Computer&13,752&767&491,722&10\\
		Amazon&Photos&7650&745&238,162&8\\
		Coauthor&CS&18,333&6805&163788&15\\
		Actor&Actor&7600&932&30019&5\\
		\noalign{\smallskip}\hline
	\end{tabular}
\end{table}

\subsection{Baselines}
We compare our proposed method, NegAmplify, with the results of thirteen SOTA self-supervised learning methods. 
These thirteen baselines have various flavors in the learning process, including negative sample-based and negative sample-free methods, data augmentation, and data augmentation-free methods, while some methods are non-contrastive methods. 
\begin{enumerate}
\item \textbf{DGI} \cite{velikovi2019deep} is a negative sample-based method that stands out as a pioneering method in graph self-supervised learning, maximizing mutual information between node representations and global graph summaries. 
\item \textbf{GRACE} \cite{DeepGrace2020} uses all nodes as negative samples and employs a double-view augmentation approach, altering both features and structure in a stochastic manner. 
\item \textbf{GCA}\cite{GCADBLP-abs-2010-14945} also leverages double-view augmentation but focuses on adaptive rather than stochastic augmentation. It uses the same negative samples and loss function as GRACE.
\item \textbf{BGRL} \cite{thakoor2021bootstrapped} similar to GRACE, uses double-view augmentation impacting both feature and structure levels. Differing from GRACE, it operates without negative samples.
\item \textbf{MVGRRL}\cite{pmlr-v119-hassani20a} is also a negative sample-based method which adopts a multi-view augmentation strategy. 
\item \textbf{LG2AR} \cite{LG2AR} is an adaptive augmentation method which also use negative samples for contrastive learning. 

\item \textbf{AFGRL}\cite{AFGCL} is an augmentation-free method and generates an alternative view of the graph by identifying nodes that share local structural information and global semantics. AFGRL method does not use negative samples in its learning process.
\item \textbf{COSTA} \cite{CostaZhang_2022} is an adaptive augmentation based method. Same like GCA and GRACE, it uses all remaining nodes as negative samples. 
\item \textbf{AF-GCL} \cite{AFGRL} doesn't require data augmentation and  utilizes graphs' inherent structure and semantic information. It selects negative sample nodes dissimilar in terms of their structural features or semantic content. 
\item  \textbf{AdaS} \cite{10181235} is an adaptive negative sampling strategy and aims to learn from the informative nodes rather than all nodes.  It also aims to learn to identify the false negative samples. 
\item  \textbf{GRAM} \cite{10025823} is a negative sample selection method that considers all nodes as negative sample candidates and then chooses negative samples from candidate nodes. 
\item \textbf{ABGML}\cite{ABGML} is an Augmentation-aware Contrastive Learning method but not a negative sampling-based method. 
\item \textbf{ProGCL} \cite{xia2022progcl} is an adaptive negative sampling-based method that enhances hard negative mining in graph contrastive learning.

\end{enumerate}

\subsection{NegAmplify Comparison with SOTA}
Table \ref{table:curriresults} presents the classification results of this method on nine benchmark datasets compared to the thirteen SOTA and three classical methods. 
The first three rows of Table \ref{table:curriresults} are not self-supervised learning but unsupervised learning methods.
Their results are not even close to being good compared to SOTA. 
However, these first three rows depict the benefits of using features and structural information on these datasets. 
\par
DGI is one of the first methods introduced in 2018. It is one of the most cited works in the graph contrastive learning field and is considered a benchmark, but its performance is outdated and is not as good as recent methods. 
NegAmplify outperforms DGI on every dataset. 
GRACE is a negative sample-based SOTA method that uses all available nodes (except the positive sample) as negative samples. 
More negative samples offer more variety to the model but increase the computation cost and overload the model with unnecessary information. 
By using half of the negative samples in NegAmplify compared to GRACE, NegAmplify outperforms GRACE on all nine datasets with up to 4.05\% classification accuracy. 
GCA, BGRL, and MVGRL had state-of-the-art results at the time of their publication, but there are methods like LG2AR that outperform BGRL on Amazon photo and Coauthor CS datasets. 
Overall, BGRL and COSTA are only two methods which uses all eight benchmark datasets except Actor dataset. 
\par
The last two rows of Table \ref{table:curriresults} present the NegAmplify results.  NegAmplify\textsubscript{c} is the CSS variant, and NegAmplify\textsubscript{r} is the random variant, which randomly selects negative samples rather than from easy, medium, and hard negative sample pools. 
NegAmplify, combined variants perform better on seven of nine datasets than all thirteen baseline methods. 
The maximum improvement in results is on the CiteSeer dataset, where the previous best results are AdaS, and NegAmplify beats them with a 2.86\% improvement. 
NegAmplify\textsubscript{c} outperforms all baselines on seven out of nine datasets, where its results on the Amazon Photo dataset are only worse than LG2AR with a marginal difference. 
The Coauthor CS dataset is also unsuitable for NegAmplify\textsubscript{c}, as its performance is not good with NegAmplify both variants. 
\par
The last row presents the NegAmplify\textsubscript{r} classification performance, randomly selecting negative samples rather than from pools. In other words, here, the quality of negative samples is not as controlled as NegAmplify\textsubscript{c}. 
NegAmplify\textsubscript{r} performs very well compared to baselines, and on six out of nine benchmark datasets, it has better results than SOTA. 
There is just a 0.02\% accuracy difference between NegAmplify\textsubscript{r} and LG2AR on the Amazon Photo dataset. 
This variant performs better than NegAmplify\textsubscript{c} on DBLP and WikiCS datasets. 
Overall, Table \ref{table:curriresults} depicts that NegAmplify is a better option than baseline methods, considering the consistent performance on multiple benchmark datasets.

\section{Negative Samples Analysis}
This section analyzes the negative samples and their impact on results. 
First, \textit{Percentage Analysis} explores the number of negative samples against each dataset to get the best classification performance. 
Second, \textit{Dataset and Negative Samples Relation} analyzes the relationship between dataset density and negative samples as dense and bigger datasets have more nodes and more possibilities of negative samples. At last, \textit{Negative Samples Hardness Analysis} presents if curriculum learning is a suitable paradigm for node classification.

\newgeometry{margin=3.5cm} 
\begin{landscape}
\begin{table*}[t]
\caption{Classification accuracy (mean $\pm$ standard deviation), where the highest results are \textbf{bold} and the second highest are \textit{italic}. We perform experiments only for \textit{NegAmplify}, while other results are reported from original papers where `-' indicates results are not in the original paper.}
\label{table:curriresults}
\begin{center}
\begin{tabular}{llllllllll}
\hline\noalign{\smallskip}
Algo &Cora &CiteSeer &Dblp & PubMed&WikiCS&Am.Comp& Am.Photo&Co.CS& Actor\\
\hline\noalign{\smallskip}
RF&64.8&64.6&71.6&84.8&71.98$\pm$0&73.81$\pm$0&78.53$\pm$0&90.37$\pm$0 & -\\
DW&75.7&50.5&75.9&80.5&74.35$\pm$0.06&85.68$\pm$0.06&89.44$\pm$0.11&84.61$\pm$0.22& -\\
DW+ft.&73.1&47.6&78.1&83.7&77.21$\pm$0.03&86.28$\pm$0.07&90.05$\pm$0.08&87.70$\pm$0.04& -\\
\hline\noalign{\smallskip}
DGI\cite{velikovi2019deep}&82.6$\pm$0.4&68.8$\pm$0.7&83.2$\pm$0.1&86.0$\pm$0.1&75.35$\pm$0.14&83.95$\pm$0.47&91.61$\pm$0.22&92.15$\pm$0.63& -\\
GRACE \cite{DeepGrace2020}&83.3$\pm$0.4&72.1$\pm$0.5& 84.2$\pm$0.1&86.7$\pm$0.1&80.14$\pm$0.48&89.53$\pm$0.35&92.78$\pm$0.45&91.12$\pm$0.20& 30.33$\pm$0.77\\
GCA\cite{GCADBLP-abs-2010-14945}&-&-& -&-&78.35$\pm$0.05&88.94$\pm$0.15&92.53$\pm$0.16&93.10$\pm$0.01& -\\
BGRL\cite{thakoor2021bootstrapped}&83.83$\pm$1.61&72.32$\pm$0.89&84.07$\pm$0.23&86.03$\pm$0.33&79.98$\pm$0.10&90.34$\pm$0.19&93.17$\pm$0.30&93.31$\pm$0.13& 27.64$\pm$0.03\\
MVGRL\cite{pmlr-v119-hassani20a}&{86.80$\pm$0.5}&73.30$\pm$0.5&-& 80.10$\pm$0.70&77.52$\pm$0.08&87.52$\pm$0.11&91.74$\pm$0.07&92.11$\pm$0.12& -\\
LG2AR\cite{LG2AR} &82.70$\pm$0.70&-&-&81.50$\pm$0.70&77.80$\pm$0.50&89.60$\pm$0.30&\textbf{94.10$\pm$0.40}&\textbf{93.60$\pm$0.30}& -\\
AFGRL\cite{AFGRL}&-&-&-&-&77.62$\pm$0.49&89.88$\pm$0.33&{93.22$\pm$0.28}&93.27$\pm$0.17& -\\
COSTA\cite{CostaZhang_2022} &84.30$\pm$0.20&72.90$\pm$0.30&84.50$\pm$0.10&86.20$\pm$0.10&79.12$\pm$0.02&88.32$\pm$0.03&92.56$\pm$0.45&92.95$\pm$0.12& -\\
AF-GCL\cite{AFGCL} &83.16$\pm$0.13&71.96$\pm$0.42&-&81.50$\pm$0.70&79.01$\pm$0.51&89.68$\pm$0.19&92.49$\pm$0.31&91.92$\pm$0.10& -\\
AdaS \cite{10181235} & 83.51$\pm$1.18  & 73.16$\pm$0.78 & -& 80.47$\pm$1.94& -& 79.10$\pm$1.65&90.63$\pm$1.13&91.63$\pm$0.57 & - \\
GRAM \cite{10025823} & 84.90$\pm$0.50  & 72.90$\pm$0.50 & {84.70$\pm$0.10}& 84.90$\pm$0.20&-& -&-&- & - \\
ABGML \cite{ABGML}  & -  & - & -& -& 78.70$\pm$0.56& 90.17$\pm$0.30&93.46$\pm$0.36&\textit{93.56$\pm$0.19} &- \\	
ProGCL\cite{xia2022progcl} &-&-&-&-&78.68$\pm$0.12&89.55$\pm$0.16&93.64$\pm$0.13&93.67$\pm$0.12& -\\			
\hline\noalign{\smallskip}
NegAmplify\textsubscript{c} &\textbf{ 87.43$\pm$0.32}& \textbf{75.92$\pm$ 0.88} & \textit{85.40$\pm$ 0.04} & \textbf{87.09$\pm$0.05} & \textit{81.69$\pm$0.10} &  \textbf{90.43$\pm$0.25}  & 93.73$\pm$0.21 & {93.15$\pm$0.07} & \textbf{31.14$\pm$0.66} \\
				
NegAmplify\textsubscript{r} &\textit{87.53$\pm$0.11}& {74.56$\pm$ 0.13} & \textbf{85.87$\pm$ 0.05} & \textit{86.46$\pm$0.16} & \textbf{82.04$\pm$0.10} &  {89.84$\pm$0.13}  & \textit{94.09$\pm$0.39} & {93.26$\pm$0.08} & \textit{30.47$\pm$0.09} \\

\noalign{\smallskip}\hline				
			\end{tabular}
		\end{center}
	\end{table*}
\end{landscape}
\restoregeometry

\subsection{Percentage Analysis}
Previous works \cite{DeepGrace2020,GCADBLP-abs-2010-14945,FebAA.2207.01792} uses all nodes as negative samples, but Table \ref{table:curriresults} proves that using half negative samples is effective rather than using all candidate nodes for negative samples. 
However, this raises a concern about how many negative samples are enough. 
In this section, we answer this question by analyzing the different percentages and present the results in Table \ref{table:percentageresults}. 
For these experiments, 0 to 100\% negative samples are used. However, the negative samples are provided to the model as per  Algorithm \ref{Algo:NegAmplify}. 
Given that, at the start, only 10\% negative samples are given as input, and later, their percentage increase as per the algorithm. 
\par
The column headings of Table \ref{table:percentageresults} present the maximum percentage of negative samples used. 
For example, 30\% indicates that 30\% negative samples are used for training while the starting value was 10\% (except 2nd column), and the decision to increase the negative sample by 1\% is dependent on Algorithm \ref{Algo:NegAmplify} (except first two columns).  
\par
Table \ref{table:percentageresults} illustrates that only using positive samples is insufficient in contrastive learning negative sample-based paradigm. 
The second column of the table presents the results where only positive samples are used for training. 
In such a situation, positive and negative samples are not contrasted; hence, these results are worst in eight of nine datasets, except the Amazon photo dataset, which is not much affected by the negative samples. 
Training with 10\% negative samples is only suitable for the Amazon photo dataset. Else all other datasets have bad results in this setting. 
\par
Overall, five out of nine datasets' best results indicate that using half of the negative samples is enough to get the best results, backed by Table \ref{table:curriresults} results. 
Cora, DBLP, and PubMed perform best when there are 80\% negative samples are used. 
Interestingly, none of the datasets have the best results on 100\% negative samples. 
Overall, the percentage of negative samples  and dataset density have direct relations where sparse datasets require more negative samples than dense datasets. 
The following section analyzes the negative samples from the dataset perspective. 

\begin{table*}[t]
	\caption{Classification accuracy (mean + standard deviation) under different (maximum) percentages of negative samples. 
	}
	\label{table:percentageresults}
	\begin{center}
		\begin{tabular}{llllllllllll}
			\hline\noalign{\smallskip}
			Dataset & 0\% & 10\% & 20\% & 30\% & 40\%& 50\% & 60\% & 70\%& 80\%& 90\% &100\% \\
			\hline\noalign{\smallskip}
			Cora & 70.58 & 84.51 & 85.66 & 86.62 & 86.84 & 87.53 & 87.46 & 87.73 & \textbf{87.90} & \textit{87.85} & 87.90 \\
			CiteSeer & 56.35 & 74.12 & 75.39 & 75.74 & 75.50 & \textbf{76.58} & \textit{76.39} & 76.10 & 76.01& 76.25 & 76.07 \\
			Dblp & 72.08 & 84.77 & 84.75 & 85.20 & 85.41 & 85.42 & 85.43 & 85.39 & \textbf{85.53}& \textit{85.49} & 85.53 \\
			PubMed & 79.16 & 86.74 & 86.69 & 86.81 & 87.00 & 87.19 & \textit{87.24} & 87.08 & \textbf{87.25}& 87.21 & 87.00 \\
			WikiCS & 81.60 & 80.89 & \textbf{82.48} & 81.94 & 82.05 & 81.95 & 82.38 & 81.73 & 82.42 & \textit{82.45} & 82.26\\
			Am. Comp & 88.95 & 90.17 & 90.32 & \textbf{90.78} & 90.54 & 90.58 & 90.51 & \textit{90.72} & 90.44 & 90.49 & 90.42\\
			Am.Photo & 94.27 & \textbf{94.39} & 94.23 & 94.17 & 94.13 & 94.13 & 94.10 & 94.33 & 94.07& 94.24 & \textit{94.35} \\
			Co.CS & 90.89 & 92.99 & \textbf{93.31} & \textit{93.31} & 93.12 & 93.19 & 93.10 & 93.05 & 93.21& 93.18 & 93.17 \\
			Actor & 30.00 & 30.21 & 30.66 & 31.66 & \textit{31.72} & 31.74 & \textbf{31.75} & 31.26 & 30.70 & 30.67 & 30.28\\
			\noalign{\smallskip}\hline
		\end{tabular}
	\end{center}
\end{table*}

\subsection{Dataset and Negative Samples Relation}
We divide the dataset into two sections based on their density: Sparse and Dense Dataset. Below, we explain the results of  Table \ref{table:percentageresults} from dataset sparsity perspective. Figure \ref{fig:c3:PercentageAnalysisA} also illustrates the results of selective datasets. 
\subsubsection{Sparse Datasets}
\begin{figure}[h]
\centering
		
\subfloat[CiteSeer]{\label{c3sfig:b}\includegraphics[width=.45\textwidth]{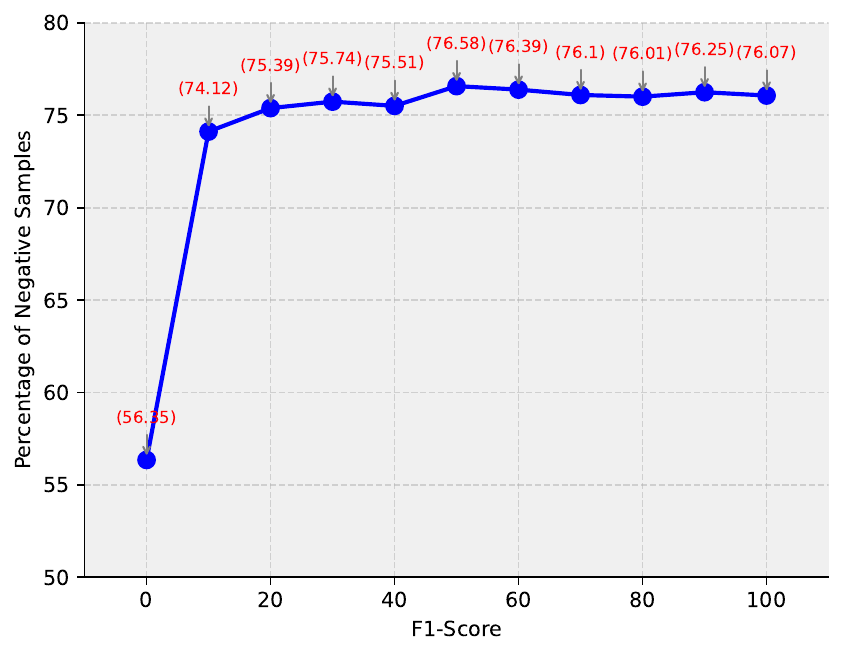}} 
\subfloat[DBLP]{\label{c3sfig:c}\includegraphics[width=.45\textwidth]{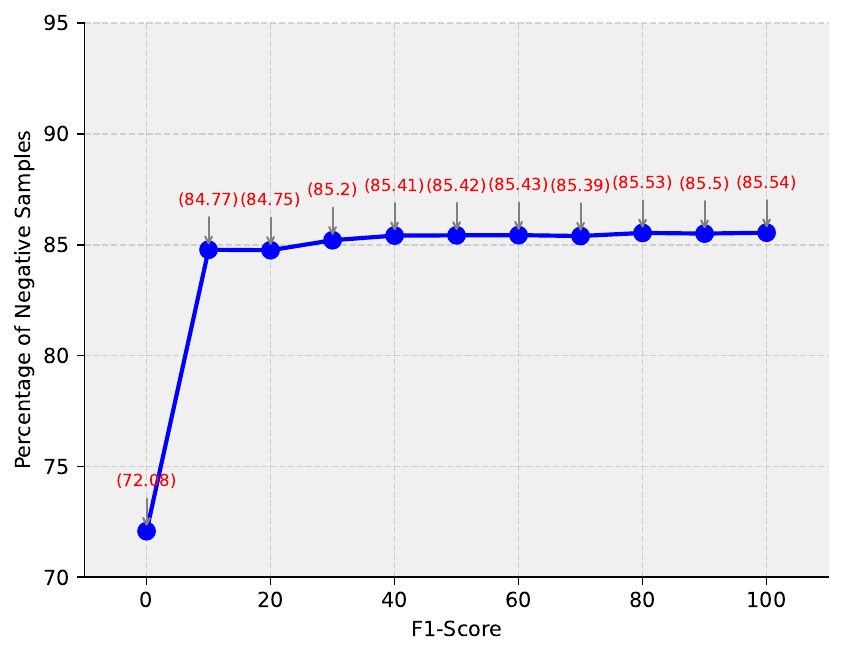}}\\
\subfloat[WikiCS]{\label{c3sfig:e}\includegraphics[width=.45\textwidth]{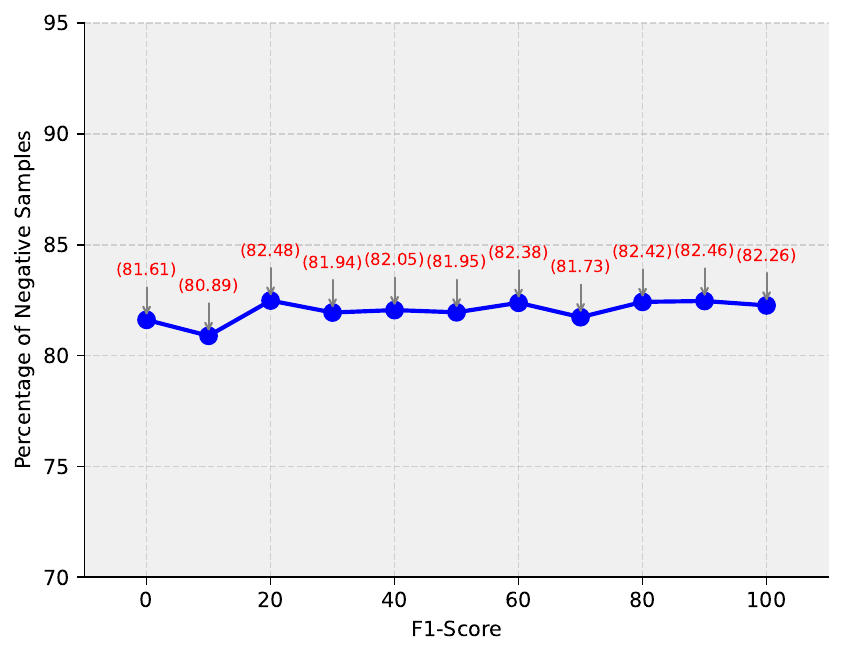}}
\subfloat[Am Photo]{\label{c3sfig:h}\includegraphics[width=.45\textwidth]{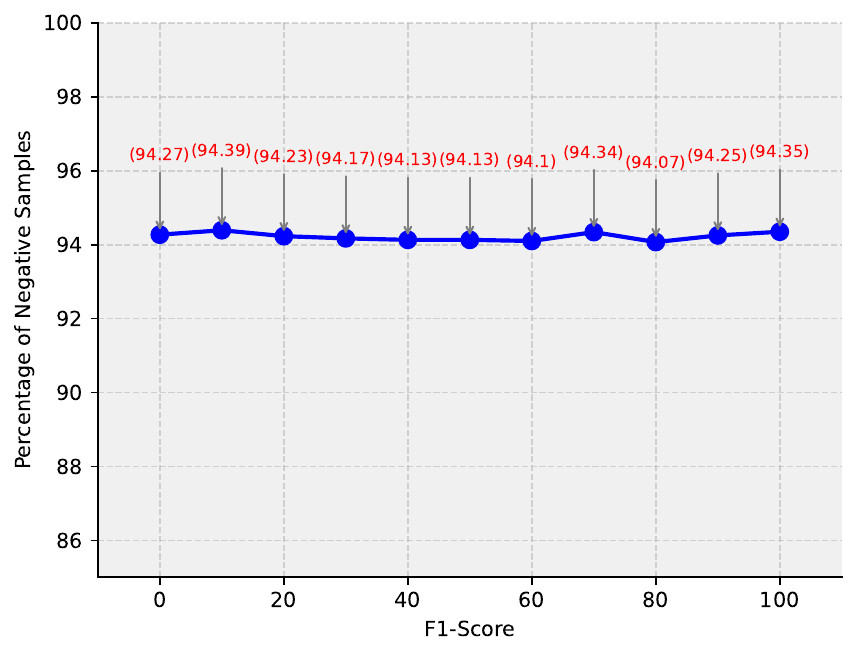}}

\caption{Influence of Negative samples on node classification performance on selective sparse and dense datasets. The results are the sum of the average and standard deviation of ten runs.} 
\label{fig:c3:PercentageAnalysisA}
\end{figure}

Table \ref{table:percentageresults} depicts the  informative behavior of different datasets, which we visualize in Figure \ref{fig:c3:PercentageAnalysisA}.
Figure \ref{c3sfig:b} presents the results of the {CiteSeer} dataset, where using no negative sample gives the worst classification results. The results on the 0\% dataset are only better than two unsupervised methods from Table \ref{table:curriresults}. 
Increasing the negative samples to 10\% has a good impact on performance, and results go up to 74.12\%, which are better than all of the baseline methods in Table  \ref{table:curriresults}. 
Increasing the negative samples from 0 to 10\% increased the classification performance to 17.77\%, which is 0.32\% better than the previously best MVGRL results.
Concluding that only 10\% negative samples beat all of the baseline methods.
Increasing the maximum percentage of negative samples gradually improves the performance, where the maximum performance we get on 50\% negative samples. 
However, after crossing the 50\% limit, classification performance decreases.
\par 
Table \ref{table:percentageresults} presents the results of the {Cora} dataset. Using zero negative samples for contrastive learning on Cora gives the worst results, which are only better from one method in Table  \ref{table:curriresults}. 
Increasing 10\% negative samples makes the accuracy to 84.52\%, which is second best after MVGRL. 
Increasing the negative samples to 50\% improves the Cora results more than all of the baselines, and improvement is very marginal (0.23\%). Table \ref{table:percentageresults} presents that the Cora dataset needs more negative samples to require the best results in contrast to CiteSeer. The best results of Cora are on 80\% negative samples. 
It is noticeable that increasing the percentage of negative samples improves Cora's performance until the best results are obtained using 80\% negative samples. 
\par
{DBLP} is a large dataset with 17,716 nodes and 105,734 edges with four classes. It is one of the least used datasets by baseline methods.
Figure \ref{c3sfig:c} illustrates the effect of the negative sample on the DBLP dataset. 
DBLP Depicts the same behavior as CiteSeer and Cora on zero negative samples and has the worst results as compared to SOTA methods.
However, increasing the negative samples to 10\% improves its classification score with 12.69\%, making it 84.77\%, which is better than all of the baselines. DBLP result behavior is identical to CORA, where increasing the negative samples increases the classification performance. 
\par

{PubMed} is also a citation dataset with 19717 nodes, 88648 edges, and just 3 classes. It has the fewest classes among all datasets, and its classification accuracy is also very stable on baseline datasets. 
On zero percent results the classification score is 79.17\%, which is worse than all baselines. However, just 10\% negative samples increase the performance to 86.74\%, making it almost as good as GRACE, while GRACE uses 100\% negative samples. 
Further increasing the number of negative samples improves the classification performance until the second-best results are achieved on 60\% of negative samples. Further improvement does not make a big difference; it improves just 0.01\% on 80\% 
\par
The {Actor} dataset is a medium size dataset with 7600 nodes and 30019 edges with 5 classes. 
Actor is not a benchmark dataset for graph contrastive learning, but we use it to increase the dataset diversity. 
The actor's classification results are not as good as those of other datasets. Its highest classification score is 31.75\%, achieved on 60\%  negative samples. 
Using no negative samples just gives a 30\% accuracy, which is just 1.75\% less than the highest results. 
Further, increasing the negative samples decreases the classification accuracy. 
\par
Concluding the results of these five datasets. Using no negative samples is ineffective for contrastive learning and did not help in learning the better embeddings. 
However, using just 10\% negative samples increases the average classification performance on all datasets, sometimes beating the majority of baselines. 
Table  \ref{table:curriresults} results also conclude that using 50\% negative samples can give better results than baselines.

\subsubsection{Dense Datasets}
Figure \ref{c3sfig:e} presents the results on the {WikiCS} dataset from 0 to 100\% negative samples. 
Starting with no negative samples, when WikiCS is trained in only positive samples, its accuracy is still better than all baselines. 
However, its performance decreases when the percentage of negative samples is 10\%, where adding another 10\% improves the performance again. 
We argue that such behavior is because WikiCS is a different kind of dataset as compared to previously discussed datasets. It has 99.99\% nonbinary numbers, while the majority of benchmark datasets are binary or one-hot datasets. It is the only dataset that has self-loops.
The highest results on the WikiCS dataset are 82.48\%, which are using only 20\% of negative samples. 
Increasing the number of negative samples does not depict a pattern or trend, and there are marginal differences in accuracy. 
\par
{Coauthor CS} is one of the large datasets with 783336 nodes and 163,788 edges, but it has 15 classes. It has less than 1\% number of ones in the feature matrix. 
Table \ref{table:percentageresults}, starting with no negative samples, gives the classification score of 90.89\%, which is worse than contrastive learning baselines. 
However, increasing the number of negative samples improves its accuracy up to 2.1\% and makes it 92.99\%. A further 10\% increase yields the best results, and the classification score goes to 93.31\%, which is better than six out of nine baselines. We could not beat the SOTA methods on this dataset with the simple settings of NegAmplify. 
Table \ref{table:percentageresults} illustrates that increasing the number of negative samples is not favorable on the Coauthor CS  dataset. 

\par
{Amazon computers} results are presented in Table \ref{table:percentageresults}. The lowest classification accuracy is 88.95\% when no negative samples are used, where increasing the negative samples increases the accuracy, and the best results are achieved on 30\% negative samples. 
The difference between the highest and lowest results on this dataset is not very large, where the lowest accuracy is 88.95 and the highest is 90.78 with 1.83\% change. 
However, the best results are better than all of the SOTA methods as presented in Table \ref{table:curriresults}.
\par
Figure \ref{c3sfig:h} presents the {Amazon photos} results.
Amazon photo data is not greatly affected by the number of negative samples, as the difference between its best and worst results is just 0.38\%. 
Using no negative samples still gives better results than eight out of nine SOTA methods. 
It should be noticed that even with zero negative samples, Amazon photos' results are not the worst performing. Still, its  results are on 60\% negative samples, which is 94.1\%, and are better than all of the SOTA methods except one. 
\par
In summary of Table \ref{table:percentageresults}, sparse datasets are more negative sample dependent as compared to dense datasets.

\subsection{Negative Samples Hardness Analysis}
CuCo \cite{cucoijcai} implemented the curriculum learning paradigm in graph level task. This section explores it for node classification. 
Is starting training with easy negative samples better, or is starting with hard or medium negative samples better? This section explores the answer to this question. 
Table \ref{tab:c3:easyvshad} compares five possible scenarios with SOTA methods. 
The first row of the table is the benchmark datasets used in this chapter for experiments. 
The second row presents the SOTA result taken from Table \ref{table:curriresults}. 
SOTA results are the best among all baselines, not just one method's results. 
\par
The third column depicts that starting with \textit{Easy} negative samples is not the best approach, and its results are only better than SOTA on only three datasets (CiteSeer, DBLP, WikiCS). 
The fourth column starts the training with medium negative samples and yields the best results in the whole table on the DBLP and WikiCS datasets, depicting that these datasets' negative middle-level samples are better than easy and hard negative samples. 
The fifth column presents the results of starting from hard negative samples, given that no easy sample was considered for these experiments. 
The setting shows the best results on two datasets, but there is no second best, given that it is only good for Amazon photo and Co-author CS datasets. 


\par
\par
The second last column of Table \ref{tab:c3:easyvshad}, its selection pool is not the same as columns 3,4 and 5. This column is randomly selected from all of the negative samples. 
It has only three \textit{second best} results, which is still better than starting with easy samples. 
Compared with SOTA, its results are better on four out of nine datasets. 
The column row presents the results following the CSS Algorithm \ref{Algo:NegAmplify}, where a combination of easy, medium, and hard negative samples are used to train the model. 
It is the best-performing variant and has the four best results in the whole table. 
However, compared with SOTA, Its results are better on seven out of nine datasets. 
	\par
Although NegAmplify\textsubscript{r} was selected from the whole pool of negative samples, there are still no best results in this column. 
We conclude and propose to follow our method for new datasets as it considers all types of negative samples. 
However, other variants may also be considered if dataset properties are known and can match the benchmark datasets.

\begin{table}[t]
		\caption{Negative samples pools comparison}
		\label{tab:c3:easyvshad}
		\begin{center}
			\begin{tabular}{lllllll}
				\hline\noalign{\smallskip}
				Dataset & SOTA & Easy & Medium & Hard & NegAmplify\textsubscript{r} & NegAmplify\textsubscript{c} \\
				\noalign{\smallskip}\hline\noalign{\smallskip}
				Cora &86.80$\pm$0.50 & 86.80$\pm$0.11 & 86.47$\pm$0.56 & 84.37$\pm$0.25 & \textit{87.53$\pm$0.11} & \textbf{87.43$\pm$0.32} \\
				CiteSeer&73.30$\pm$0.50 & 73.74$\pm$0.36 & 73.08$\pm$0.09 & 31.11$\pm$0.37 & \textit{74.56$\pm$ 0.13} &    \textbf{75.92$\pm$ 0.88}\\
				Dblp&84.50$\pm$0.10 & \textit{86.44$\pm$0.08} & \textbf{86.81$\pm$0.06} & 85.01$\pm$0.13 & {85.87$\pm$ 0.05} & {85.40$\pm$ 0.04} \\
				PubMed&86.70$\pm$0.70& \textit{86.94$\pm$0.06} & 86.82$\pm$0.06 & 86.56$\pm$0.09 & {86.46$\pm$0.16} & \textbf{87.09$\pm$0.05} \\
				\noalign{\smallskip}\hline\noalign{\smallskip}
				WikiCS &80.14$\pm$0.48 & \textit{82.17$\pm$0.28} & \textbf{82.50$\pm$0.09} & 82.14$\pm$0.15 & {82.04$\pm$0.10} & {81.69$\pm$0.10} \\
				Am. Comp &\textit{90.34$\pm$0.19} & {89.99$\pm$0.28} & 89.91$\pm$0.21 & 90.12$\pm$0.24 & {89.84$\pm$0.13} & \textbf{90.43$\pm$0.25} \\
				Am. Photo &94.10$\pm$0.40 & 94.09$\pm$0.34 & 94.12$\pm$0.13 & \textbf{95.20$\pm$0.18} & \textit{94.09$\pm$0.39} & 93.73$\pm$0.21 \\
				Co.CS &\textit{93.60$\pm$0.30}& 92.88$\pm$0.06 & 92.81$\pm$0.02 & \textbf{93.74$\pm$0.05} & {93.26$\pm$0.08} & {93.15$\pm$0.07} \\
				Actor &30.33$\pm$0.77& 30.28$\pm$0.09 & \textit{30.93$\pm$0.22} & 30.39$\pm$0.63 & {30.47$\pm$0.09}  & \textbf{31.14$\pm$0.66} \\
				\noalign{\smallskip}\hline
				
			\end{tabular}
		\end{center}
\end{table}

\section{Conclusion}
\label{sec:c3Summary}
This study proposes a novel negative sample selection method: \textit{Cumulative Sample Selection (CSS)} and integrates it in a GCL method called ``\textit{NegAmplify}: Conditional Increase in Negative Samples".
NegAmplify aims to enhance the learning efficiency and accuracy of GCL models by optimizing the selection of negative samples, a critical yet often understated aspect of the contrastive learning paradigm.
NegAmplify controls the quality of negative samples by dividing them into three pools: easy, medium, and hard negative samples.
CSS uses the negative samples from these pools and controls the quantity and variety of negative samples. 
It starts training the model with 10\% negative samples and randomly changes the negative samples at every epoch to offer variety while controlling the quantity. A decision agent in CSS increases the negative sample quantity based on a list of previous contrastive losses. 
NegAmplify improved the classification accuracy of seven of nine datasets up to 2.86\%. 
We further investigate the impact of percentages of negative samples and conclude that sparse datasets require more negative samples than dense datasets to get better results.
We compare the pools, starting with easy, medium, and hard negative samples, to analyze which approach is better, starting with easy, medium, hard, or as proposed in CSS. 
Results show that the CSS selection method performs better than all other alternatives. 



\section*{Declaration of interests}
\noindent The authors declare that they have no known competing financial interests or personal relationships that could have appeared to influence the work reported in this paper.

\section*{CRediT author statement}
\textbf{Adnan Ali:}  Writing- original draft, Conceptualization, Methodology, Software, Visualization. \textbf{Jinlong Li, Huanhuan Chen}: Validation, Supervision, Investigation, Writing- review and editing. \textbf{Ali Kashif Bashir}: Validation, Writing- review and editing.

\section*{Acknowledgments}
The first author expresses gratitude for the collaborative PhD fellowship awarded by the Chinese Academy of Sciences (CAS) and The World Academy of Sciences (TWAS) as the first author was supported by The CAS-TWAS President's Fellowship for PhD at the University of Science and Technology China. This paper was produced during his PhD at the University of Science and Technology China.

\bibliographystyle{elsarticle-num} 
\bibliography{bib}

\begin{thebibliography}{10}
\expandafter\ifx\csname url\endcsname\relax
  \def\url#1{\texttt{#1}}\fi
\expandafter\ifx\csname urlprefix\endcsname\relax\def\urlprefix{URL }\fi
\expandafter\ifx\csname href\endcsname\relax
  \def\href#1#2{#2} \def\path#1{#1}\fi

\bibitem{ZHANG2024111558}
Y.~Zhang, J.~Zhu, Y.~Zhang, Y.~Zhu, J.~Zhou, Y.~Xie, \href{https://www.sciencedirect.com/science/article/pii/S1568494624003326}{Social-aware graph contrastive learning for recommender systems}, Applied Soft Computing 158 (2024) 111558.
\newblock \href {https://doi.org/https://doi.org/10.1016/j.asoc.2024.111558} {\path{doi:https://doi.org/10.1016/j.asoc.2024.111558}}.
\newline\urlprefix\url{https://www.sciencedirect.com/science/article/pii/S1568494624003326}

\bibitem{DBLP:journals/corr/abs-2102-10757}
Y.~Xie, Z.~Xu, Z.~Wang, S.~Ji, \href{https://arxiv.org/abs/2102.10757}{Self-supervised learning of graph neural networks: {A} unified review}, CoRR abs/2102.10757 (2021).
\newblock \href {http://arxiv.org/abs/2102.10757} {\path{arXiv:2102.10757}}.
\newline\urlprefix\url{https://arxiv.org/abs/2102.10757}

\bibitem{BookHamilton}
W.~L. Hamilton, Graph representation learning, Synthesis Lectures on Artificial Intelligence and Machine Learning 14~(3) (2020) 1--159.

\bibitem{MA2023110388}
S.~Ma, J.~wei Liu, X.~Zuo, \href{https://www.sciencedirect.com/science/article/pii/S1568494623004064}{Self-supervised learning for heterogeneous graph via structure information based on metapath}, Applied Soft Computing 143 (2023) 110388.
\newblock \href {https://doi.org/https://doi.org/10.1016/j.asoc.2023.110388} {\path{doi:https://doi.org/10.1016/j.asoc.2023.110388}}.
\newline\urlprefix\url{https://www.sciencedirect.com/science/article/pii/S1568494623004064}

\bibitem{GraphSageHamiltonYL17}
W.~L. Hamilton, R.~Ying, J.~Leskovec, \href{http://arxiv.org/abs/1706.02216}{Inductive representation learning on large graphs}, CoRR abs/1706.02216 (2017).
\newblock \href {http://arxiv.org/abs/1706.02216} {\path{arXiv:1706.02216}}.
\newline\urlprefix\url{http://arxiv.org/abs/1706.02216}

\bibitem{KipfW16}
T.~N. Kipf, M.~Welling, \href{http://arxiv.org/abs/1609.02907}{Semi-supervised classification with graph convolutional networks}, CoRR abs/1609.02907 (2016).
\newblock \href {http://arxiv.org/abs/1609.02907} {\path{arXiv:1609.02907}}.
\newline\urlprefix\url{http://arxiv.org/abs/1609.02907}

\bibitem{velickovic2018graph}
P.~Veli{\v{c}}kovi{\'{c}}, G.~Cucurull, A.~Casanova, A.~Romero, P.~Li{\`{o}}, Y.~Bengio, \href{https://openreview.net/forum?id=rJXMpikCZ}{{Graph Attention Networks}}, International Conference on Learning RepresentationsAccepted as poster (2018).
\newline\urlprefix\url{https://openreview.net/forum?id=rJXMpikCZ}

\bibitem{WANG2024111512}
S.~Wang, Y.~Zhang, X.~Piao, X.~Lin, Y.~Hu, B.~Yin, \href{https://www.sciencedirect.com/science/article/pii/S1568494624002862}{Data-unbalanced traffic accident prediction via adaptive graph and self-supervised learning}, Applied Soft Computing 157 (2024) 111512.
\newblock \href {https://doi.org/https://doi.org/10.1016/j.asoc.2024.111512} {\path{doi:https://doi.org/10.1016/j.asoc.2024.111512}}.
\newline\urlprefix\url{https://www.sciencedirect.com/science/article/pii/S1568494624002862}

\bibitem{graphsofdoi.13964}
Z.~Ning, P.~Wang, P.~Wang, Z.~Qiao, W.~Fan, D.~Zhang, Y.~Du, Y.~Zhou, \href{https://arxiv.org/abs/2209.13964}{Graph soft-contrastive learning via neighborhood ranking} (2022).
\newblock \href {https://doi.org/10.48550/ARXIV.2209.13964} {\path{doi:10.48550/ARXIV.2209.13964}}.
\newline\urlprefix\url{https://arxiv.org/abs/2209.13964}

\bibitem{DBLP:journals/corr/abs-2006-10141}
W.~Jin, T.~Derr, H.~Liu, Y.~Wang, S.~Wang, Z.~Liu, J.~Tang, \href{https://arxiv.org/abs/2006.10141}{Self-supervised learning on graphs: Deep insights and new direction}, CoRR abs/2006.10141 (2020).
\newblock \href {http://arxiv.org/abs/2006.10141} {\path{arXiv:2006.10141}}.
\newline\urlprefix\url{https://arxiv.org/abs/2006.10141}

\bibitem{HGMAE2023}
Y.~Tian, K.~Dong, C.~Zhang, C.~Zhang, N.~V. Chawla, \href{https://doi.org/10.1609/aaai.v37i8.26192}{Heterogeneous graph masked autoencoders}, in: Proceedings of the Thirty-Seventh AAAI Conference on Artificial Intelligence and Thirty-Fifth Conference on Innovative Applications of Artificial Intelligence and Thirteenth Symposium on Educational Advances in Artificial Intelligence, AAAI'23/IAAI'23/EAAI'23, AAAI Press, 2023.
\newblock \href {https://doi.org/10.1609/aaai.v37i8.26192} {\path{doi:10.1609/aaai.v37i8.26192}}.
\newline\urlprefix\url{https://doi.org/10.1609/aaai.v37i8.26192}

\bibitem{LIANG2023156}
H.~Liang, X.~Du, B.~Zhu, Z.~Ma, K.~Chen, J.~Gao, \href{https://www.sciencedirect.com/science/article/pii/S0893608023001788}{Graph contrastive learning with implicit augmentations}, Neural Networks 163 (2023) 156--164.
\newblock \href {https://doi.org/https://doi.org/10.1016/j.neunet.2023.04.001} {\path{doi:https://doi.org/10.1016/j.neunet.2023.04.001}}.
\newline\urlprefix\url{https://www.sciencedirect.com/science/article/pii/S0893608023001788}

\bibitem{10025823}
C.-Y. Zhang, H.-C. Cai, C.~L.~P. Chen, Y.-N. Lin, W.-P. Fang, Graph representation learning with adaptive metric, IEEE Transactions on Network Science and Engineering 10~(4) (2023) 2074--2085.
\newblock \href {https://doi.org/10.1109/TNSE.2023.3239661} {\path{doi:10.1109/TNSE.2023.3239661}}.

\bibitem{GCADBLP-abs-2010-14945}
Y.~Zhu, Y.~Xu, F.~Yu, Q.~Liu, S.~Wu, L.~Wang, \href{https://doi.org/10.1145/3442381.3449802}{Graph contrastive learning with adaptive augmentation}, in: J.~Leskovec, M.~Grobelnik, M.~Najork, J.~Tang, L.~Zia (Eds.), {WWW} '21: The Web Conference 2021, Virtual Event / Ljubljana, Slovenia, April 19-23, 2021, {ACM} / {IW3C2}, 2021, pp. 2069--2080.
\newblock \href {https://doi.org/10.1145/3442381.3449802} {\path{doi:10.1145/3442381.3449802}}.
\newline\urlprefix\url{https://doi.org/10.1145/3442381.3449802}

\bibitem{thakoor2021bootstrapped}
S.~Thakoor, C.~Tallec, M.~G. Azar, M.~Azabou, E.~L. Dyer, R.~Munos, P.~Veličković, M.~Valko, Large-scale representation learning on graphs via bootstrapping (2021).
\newblock \href {http://arxiv.org/abs/2102.06514} {\path{arXiv:2102.06514}}.

\bibitem{DeepGrace2020}
Y.~Zhu, Y.~Xu, F.~Yu, Q.~Liu, S.~Wu, L.~Wang, \href{https://arxiv.org/abs/2006.04131}{Deep graph contrastive representation learning}, CoRR abs/2006.04131 (2020).
\newblock \href {http://arxiv.org/abs/2006.04131} {\path{arXiv:2006.04131}}.
\newline\urlprefix\url{https://arxiv.org/abs/2006.04131}

\bibitem{DAENS}
A.~Ali, J.~Li, H.~Chen, \href{https://www.sciencedirect.com/science/article/pii/S0045790624002325}{Heterogeneous data augmentation in graph contrastive learning for effective negative samples}, Computers and Electrical Engineering 118 (2024) 109304.
\newblock \href {https://doi.org/https://doi.org/10.1016/j.compeleceng.2024.109304} {\path{doi:https://doi.org/10.1016/j.compeleceng.2024.109304}}.
\newline\urlprefix\url{https://www.sciencedirect.com/science/article/pii/S0045790624002325}

\bibitem{pmlr-v198-zhao22a}
T.~Zhao, X.~Tang, D.~Zhang, H.~Jiang, N.~Rao, Y.~Song, P.~Agrawal, K.~Subbian, B.~Yin, M.~Jiang, \href{https://proceedings.mlr.press/v198/zhao22a.html}{Autogda: Automated graph data augmentation for node classification}, in: B.~Rieck, R.~Pascanu (Eds.), Proceedings of the First Learning on Graphs Conference, Vol. 198 of Proceedings of Machine Learning Research, PMLR, 2022, pp. 32:1--32:17.
\newline\urlprefix\url{https://proceedings.mlr.press/v198/zhao22a.html}

\bibitem{comunityGuo2023}
K.~Guo, J.~Lin, Q.~Zhuang, R.~Zeng, J.~Wang, \href{https://doi.org/10.1007/s10489-023-05046-w}{Adaptive graph contrastive learning for community detection}, Applied Intelligence 53~(23) (2023) 28768--28786.
\newblock \href {https://doi.org/10.1007/s10489-023-05046-w} {\path{doi:10.1007/s10489-023-05046-w}}.
\newline\urlprefix\url{https://doi.org/10.1007/s10489-023-05046-w}

\bibitem{10095350}
Y.~Zhang, H.~Zhu, S.~Yu, Adaptive data augmentation for contrastive learning, in: ICASSP 2023 - 2023 IEEE International Conference on Acoustics, Speech and Signal Processing (ICASSP), 2023, pp. 1--5.
\newblock \href {https://doi.org/10.1109/ICASSP49357.2023.10095350} {\path{doi:10.1109/ICASSP49357.2023.10095350}}.

\bibitem{FebAA.2207.01792}
A.~Ali, J.~Li, \href{https://www.sciencedirect.com/science/article/pii/S1051200423004074}{Features based adaptive augmentation for graph contrastive learning}, Digital Signal Processing (2023) 104312\href {https://doi.org/https://doi.org/10.1016/j.dsp.2023.104312} {\path{doi:https://doi.org/10.1016/j.dsp.2023.104312}}.
\newline\urlprefix\url{https://www.sciencedirect.com/science/article/pii/S1051200423004074}

\bibitem{AFGRL}
N.~Lee, J.~Lee, C.~Park, \href{https://arxiv.org/abs/2112.02472}{Augmentation-free self-supervised learning on graphs}, CoRR abs/2112.02472 (2021).
\newblock \href {http://arxiv.org/abs/2112.02472} {\path{arXiv:2112.02472}}.
\newline\urlprefix\url{https://arxiv.org/abs/2112.02472}

\bibitem{AFGCL}
H.~Wang, J.~Zhang, Q.~Zhu, W.~Huang, \href{https://arxiv.org/abs/2204.04874}{Augmentation-free graph contrastive learning with performance guarantee} (2022).
\newblock \href {https://doi.org/10.48550/ARXIV.2204.04874} {\path{doi:10.48550/ARXIV.2204.04874}}.
\newline\urlprefix\url{https://arxiv.org/abs/2204.04874}

\bibitem{cucoijcai}
G.~Chu, X.~Wang, C.~Shi, X.~Jiang, \href{https://doi.org/10.24963/ijcai.2021/317}{Cuco: Graph representation with curriculum contrastive learning}, in: Z.-H. Zhou (Ed.), Proceedings of the Thirtieth International Joint Conference on Artificial Intelligence, {IJCAI-21}, International Joint Conferences on Artificial Intelligence Organization, 2021, pp. 2300--2306, main Track.
\newblock \href {https://doi.org/10.24963/ijcai.2021/317} {\path{doi:10.24963/ijcai.2021/317}}.
\newline\urlprefix\url{https://doi.org/10.24963/ijcai.2021/317}

\bibitem{MIAO2022667}
R.~Miao, Y.~Yang, Y.~Ma, X.~Juan, H.~Xue, J.~Tang, Y.~Wang, X.~Wang, \href{https://www.sciencedirect.com/science/article/pii/S0020025522010684}{Negative samples selecting strategy for graph contrastive learning}, Information Sciences 613 (2022) 667--681.
\newblock \href {https://doi.org/https://doi.org/10.1016/j.ins.2022.09.024} {\path{doi:https://doi.org/10.1016/j.ins.2022.09.024}}.
\newline\urlprefix\url{https://www.sciencedirect.com/science/article/pii/S0020025522010684}

\bibitem{Yang_2023}
H.~Yang, H.~Chen, S.~Zhang, X.~Sun, Q.~Li, X.~Zhao, G.~Xu, \href{http://dx.doi.org/10.1145/3543507.3583499}{Generating counterfactual hard negative samples for graph contrastive learning}, in: Proceedings of the ACM Web Conference 2023, WWW ’23, ACM, 2023.
\newblock \href {https://doi.org/10.1145/3543507.3583499} {\path{doi:10.1145/3543507.3583499}}.
\newline\urlprefix\url{http://dx.doi.org/10.1145/3543507.3583499}

\bibitem{10181235}
S.~Wan, Y.~Zhan, S.~Chen, S.~Pan, J.~Yang, D.~Tao, C.~Gong, Boosting graph contrastive learning via adaptive sampling, IEEE Transactions on Neural Networks and Learning Systems (2023) 1--13\href {https://doi.org/10.1109/TNNLS.2023.3291358} {\path{doi:10.1109/TNNLS.2023.3291358}}.

\bibitem{hal-03575619}
H.~Hafidi, M.~Ghogho, P.~Ciblat, A.~Swami, \href{https://www.sciencedirect.com/science/article/pii/S0165168421003479}{Negative sampling strategies for contrastive self-supervised learning of graph representations}, Signal Processing 190 (2022) 108310.
\newblock \href {https://doi.org/https://doi.org/10.1016/j.sigpro.2021.108310} {\path{doi:https://doi.org/10.1016/j.sigpro.2021.108310}}.
\newline\urlprefix\url{https://www.sciencedirect.com/science/article/pii/S0165168421003479}

\bibitem{10382709}
C.~Niu, G.~Pang, L.~Chen, Affinity uncertainty-based hard negative mining in graph contrastive learning, IEEE Transactions on Neural Networks and Learning Systems (2024) 1--11\href {https://doi.org/10.1109/TNNLS.2023.3339770} {\path{doi:10.1109/TNNLS.2023.3339770}}.

\bibitem{Zhang2024}
B.~Zhang, L.~Wang, \href{https://www.sciopen.com/article/10.26599/TST.2023.9010043}{False negative sample detection for graph contrastive learning}, Tsinghua Science and Technology 29~(2) (2024) 529--542.
\newblock \href {https://doi.org/10.26599/TST.2023.9010043} {\path{doi:10.26599/TST.2023.9010043}}.
\newline\urlprefix\url{https://www.sciopen.com/article/10.26599/TST.2023.9010043}

\bibitem{ABGML}
D.~Chen, X.~Zhao, W.~Wang, Z.~Tan, W.~Xiao, \href{https://doi.org/10.1145/3543507.3583246}{Graph self-supervised learning with augmentation-aware contrastive learning}, in: Proceedings of the ACM Web Conference 2023, WWW '23, Association for Computing Machinery, New York, NY, USA, 2023, p. 154–164.
\newblock \href {https://doi.org/10.1145/3543507.3583246} {\path{doi:10.1145/3543507.3583246}}.
\newline\urlprefix\url{https://doi.org/10.1145/3543507.3583246}

\bibitem{word2vec}
T.~Mikolov, K.~Chen, G.~Corrado, J.~Dean, Efficient estimation of word representations in vector space (2013).
\newblock \href {http://arxiv.org/abs/1301.3781} {\path{arXiv:1301.3781}}.

\bibitem{DBLP:journals/corr/abs-2002-05709}
T.~Chen, S.~Kornblith, M.~Norouzi, G.~E. Hinton, \href{https://arxiv.org/abs/2002.05709}{A simple framework for contrastive learning of visual representations}, CoRR abs/2002.05709 (2020).
\newblock \href {http://arxiv.org/abs/2002.05709} {\path{arXiv:2002.05709}}.
\newline\urlprefix\url{https://arxiv.org/abs/2002.05709}

\bibitem{ContrastiveGenerative}
L.~Wu, H.~Lin, C.~Tan, Z.~Gao, S.~Z. Li, \href{https://doi.org/10.1109/TKDE.2021.3131584}{Self-supervised learning on graphs: Contrastive, generative, or predictive}, IEEE Trans. on Knowl. and Data Eng. 35~(4) (2023) 4216–4235.
\newblock \href {https://doi.org/10.1109/TKDE.2021.3131584} {\path{doi:10.1109/TKDE.2021.3131584}}.
\newline\urlprefix\url{https://doi.org/10.1109/TKDE.2021.3131584}

\bibitem{Deepwalk32}
B.~Perozzi, R.~Al-Rfou, S.~Skiena, \href{https://doi.org/10.1145/2623330.2623732}{Deepwalk: online learning of social representations}, in: Proceedings of the 20th ACM SIGKDD International Conference on Knowledge Discovery and Data Mining, KDD '14, Association for Computing Machinery, New York, NY, USA, 2014, p. 701–710.
\newblock \href {https://doi.org/10.1145/2623330.2623732} {\path{doi:10.1145/2623330.2623732}}.
\newline\urlprefix\url{https://doi.org/10.1145/2623330.2623732}

\bibitem{negativeReview}
L.~Xu, J.~Lian, W.~X. Zhao, M.~Gong, L.~Shou, D.~Jiang, X.~Xie, J.-R. Wen, Negative sampling for contrastive representation learning: A review (2022).
\newblock \href {http://arxiv.org/abs/2206.00212} {\path{arXiv:2206.00212}}.

\bibitem{HardNegativerobinson2021}
J.~D. Robinson, C.-Y. Chuang, S.~Sra, S.~Jegelka, \href{https://openreview.net/forum?id=CR1XOQ0UTh-}{Contrastive learning with hard negative samples}, in: International Conference on Learning Representations, 2021.
\newline\urlprefix\url{https://openreview.net/forum?id=CR1XOQ0UTh-}

\bibitem{node2vec}
A.~Grover, J.~Leskovec, \href{https://doi.org/10.1145/2939672.2939754}{node2vec: Scalable feature learning for networks}, in: Proceedings of the 22nd ACM SIGKDD International Conference on Knowledge Discovery and Data Mining, KDD '16, Association for Computing Machinery, New York, NY, USA, 2016, p. 855–864.
\newblock \href {https://doi.org/10.1145/2939672.2939754} {\path{doi:10.1145/2939672.2939754}}.
\newline\urlprefix\url{https://doi.org/10.1145/2939672.2939754}

\bibitem{xia2022progcl}
J.~Xia, L.~Wu, G.~Wang, S.~Z. Li, Progcl: Rethinking hard negative mining in graph contrastive learning, in: International conference on machine learning, PMLR, 2022.

\bibitem{zhao2021graph}
H.~Zhao, X.~Yang, Z.~Wang, E.~Yang, C.~Deng, Graph debiased contrastive learning with joint representation clustering, in: Proc. IJCAI, 2021, pp. 3434--3440.

\bibitem{CostaZhang_2022}
Y.~Zhang, H.~Zhu, Z.~Song, P.~Koniusz, I.~King, \href{https://doi.org/10.1145/3534678.3539425}{Costa: Covariance-preserving feature augmentation for graph contrastive learning}, in: Proceedings of the 28th ACM SIGKDD Conference on Knowledge Discovery and Data Mining, KDD '22, Association for Computing Machinery, New York, NY, USA, 2022, p. 2524–2534.
\newblock \href {https://doi.org/10.1145/3534678.3539425} {\path{doi:10.1145/3534678.3539425}}.
\newline\urlprefix\url{https://doi.org/10.1145/3534678.3539425}

\bibitem{zhu2021empirical}
Y.~Zhu, Y.~Xu, Q.~Liu, S.~Wu, \href{https://arxiv.org/abs/2109.01116}{An empirical study of graph contrastive learning}, CoRR abs/2109.01116 (2021).
\newblock \href {http://arxiv.org/abs/2109.01116} {\path{arXiv:2109.01116}}.
\newline\urlprefix\url{https://arxiv.org/abs/2109.01116}

\bibitem{Schroff_2015_CVPR}
F.~Schroff, D.~Kalenichenko, J.~Philbin, Facenet: A unified embedding for face recognition and clustering, in: Proceedings of the IEEE Conference on Computer Vision and Pattern Recognition (CVPR), 2015.

\bibitem{s23125572}
X.-Y. Yang, F.~Xu, J.~Yu, Z.-Y. Li, D.-X. Wang, \href{https://www.mdpi.com/1424-8220/23/12/5572}{Graph neural network-guided contrastive learning for sequential recommendation}, Sensors 23~(12) (2023).
\newblock \href {https://doi.org/10.3390/s23125572} {\path{doi:10.3390/s23125572}}.
\newline\urlprefix\url{https://www.mdpi.com/1424-8220/23/12/5572}

\bibitem{liu2023hard}
Y.~Liu, X.~Yang, S.~Zhou, X.~Liu, Z.~Wang, K.~Liang, W.~Tu, L.~Li, J.~Duan, C.~Chen, Hard sample aware network for contrastive deep graph clustering (2023).
\newblock \href {http://arxiv.org/abs/2212.08665} {\path{arXiv:2212.08665}}.

\bibitem{8578392}
Y.~Duan, W.~Zheng, X.~Lin, J.~Lu, J.~Zhou, Deep adversarial metric learning, in: 2018 IEEE/CVF Conference on Computer Vision and Pattern Recognition, 2018, pp. 2780--2789.
\newblock \href {https://doi.org/10.1109/CVPR.2018.00294} {\path{doi:10.1109/CVPR.2018.00294}}.

\bibitem{DBLP:journals/corr/abs-1903-05503}
W.~Zheng, Z.~Chen, J.~Lu, J.~Zhou, \href{http://arxiv.org/abs/1903.05503}{Hardness-aware deep metric learning}, CoRR abs/1903.05503 (2019).
\newblock \href {http://arxiv.org/abs/1903.05503} {\path{arXiv:1903.05503}}.
\newline\urlprefix\url{http://arxiv.org/abs/1903.05503}

\bibitem{Fan_2023}
L.~Fan, J.~Pu, R.~Zhang, X.-M. Wu, \href{http://dx.doi.org/10.1145/3539618.3591995}{Neighborhood-based hard negative mining for sequential recommendation}, in: Proceedings of the 46th International ACM SIGIR Conference on Research and Development in Information Retrieval, SIGIR ’23, ACM, 2023.
\newblock \href {https://doi.org/10.1145/3539618.3591995} {\path{doi:10.1145/3539618.3591995}}.
\newline\urlprefix\url{http://dx.doi.org/10.1145/3539618.3591995}

\bibitem{DBLP:journals/corr/abs-1811-04155}
D.~H. Park, Y.~Chang, \href{http://arxiv.org/abs/1811.04155}{Adversarial sampling and training for semi-supervised information retrieval}, CoRR abs/1811.04155 (2018).
\newblock \href {http://arxiv.org/abs/1811.04155} {\path{arXiv:1811.04155}}.
\newline\urlprefix\url{http://arxiv.org/abs/1811.04155}

\bibitem{feng2022adversarial}
S.~Feng, B.~Jing, Y.~Zhu, H.~Tong, \href{https://doi.org/10.1145/3485447.3512183}{Adversarial graph contrastive learning with information regularization}, in: Proceedings of the ACM Web Conference 2022, WWW '22, Association for Computing Machinery, New York, NY, USA, 2022, p. 1362–1371.
\newblock \href {https://doi.org/10.1145/3485447.3512183} {\path{doi:10.1145/3485447.3512183}}.
\newline\urlprefix\url{https://doi.org/10.1145/3485447.3512183}

\bibitem{velikovi2019deep}
P.~Veličković, W.~Fedus, W.~L. Hamilton, P.~Liò, Y.~Bengio, D.~Hjelm, \href{https://www.microsoft.com/en-us/research/publication/deep-graph-infomax/}{Deep graph infomax}, in: ICLR 2019, 2019.
\newline\urlprefix\url{https://www.microsoft.com/en-us/research/publication/deep-graph-infomax/}

\bibitem{CiteseerDataset}
C.~L. Giles, K.~D. Bollacker, S.~Lawrence, \href{https://doi.org/10.1145/276675.276685}{Citeseer: an automatic citation indexing system}, in: Proceedings of the Third ACM Conference on Digital Libraries, DL '98, Association for Computing Machinery, New York, NY, USA, 1998, p. 89–98.
\newblock \href {https://doi.org/10.1145/276675.276685} {\path{doi:10.1145/276675.276685}}.
\newline\urlprefix\url{https://doi.org/10.1145/276675.276685}

\bibitem{CoraDataset}
A.~K. McCallum, K.~Nigam, J.~Rennie, K.~Seymore, \href{https://doi.org/10.1023/A:1009953814988}{Automating the construction of internet portals with machine learning}, Information Retrieval 3~(2) (2000) 127--163.
\newblock \href {https://doi.org/10.1023/A:1009953814988} {\path{doi:10.1023/A:1009953814988}}.
\newline\urlprefix\url{https://doi.org/10.1023/A:1009953814988}

\bibitem{PubMedDataset}
P.~Sen, G.~Namata, M.~Bilgic, L.~Getoor, B.~Galligher, T.~Eliassi-Rad, \href{https://ojs.aaai.org/aimagazine/index.php/aimagazine/article/view/2157}{Collective classification in network data}, AI Magazine 29~(3) (2008) 93.
\newblock \href {https://doi.org/10.1609/aimag.v29i3.2157} {\path{doi:10.1609/aimag.v29i3.2157}}.
\newline\urlprefix\url{https://ojs.aaai.org/aimagazine/index.php/aimagazine/article/view/2157}

\bibitem{dblpdataset}
A.~Bojchevski, S.~Günnemann, Deep gaussian embedding of graphs: Unsupervised inductive learning via ranking (2018).
\newblock \href {http://arxiv.org/abs/1707.03815} {\path{arXiv:1707.03815}}.

\bibitem{pmlr-v119-hassani20a}
K.~Hassani, A.~H. Khasahmadi, \href{https://proceedings.mlr.press/v119/hassani20a.html}{Contrastive multi-view representation learning on graphs}, in: H.~D. III, A.~Singh (Eds.), Proceedings of the 37th International Conference on Machine Learning, Vol. 119 of Proceedings of Machine Learning Research, PMLR, 2020, pp. 4116--4126.
\newline\urlprefix\url{https://proceedings.mlr.press/v119/hassani20a.html}

\bibitem{LG2AR}
K.~Hassani, A.~H.~K. Ahmadi, \href{https://arxiv.org/abs/2201.09830}{Learning graph augmentations to learn graph representations}, CoRR abs/2201.09830 (2022).
\newblock \href {http://arxiv.org/abs/2201.09830} {\path{arXiv:2201.09830}}.
\newline\urlprefix\url{https://arxiv.org/abs/2201.09830}

\bibitem{2019arXiv190302428F}
M.~{Fey}, J.~E. {Lenssen}, {Fast Graph Representation Learning with PyTorch Geometric}, arXiv e-prints (2019) arXiv:1903.02428\href {http://arxiv.org/abs/1903.02428} {\path{arXiv:1903.02428}}.

\bibitem{NEURIPS2019_bdbca288}
A.~Paszke, S.~Gross, F.~Massa, A.~Lerer, J.~Bradbury, G.~Chanan, T.~Killeen, Z.~Lin, N.~Gimelshein, L.~Antiga, A.~Desmaison, A.~Kopf, E.~Yang, Z.~DeVito, M.~Raison, A.~Tejani, S.~Chilamkurthy, B.~Steiner, L.~Fang, J.~Bai, S.~Chintala, \href{https://proceedings.neurips.cc/paper/2019/file/bdbca288fee7f92f2bfa9f7012727740-Paper.pdf}{Pytorch: An imperative style, high-performance deep learning library}, in: H.~Wallach, H.~Larochelle, A.~Beygelzimer, F.~d\textquotesingle Alch\'{e}-Buc, E.~Fox, R.~Garnett (Eds.), Advances in Neural Information Processing Systems, Vol.~32, Curran Associates, Inc., 2019.
\newline\urlprefix\url{https://proceedings.neurips.cc/paper/2019/file/bdbca288fee7f92f2bfa9f7012727740-Paper.pdf}

\end{thebibliography}

\appendix

\section{Implementation Details}
\label{sec:c3:implementation}
This section presents the \textit{Software, Hardware}, and \textit{Hyper-parameter} details to regenerate the results, where implementation of NegAmplify is available on the GitHub link: https://github.com//mhadnanali/NegAmplify.
\subsection{Software and Hardware Details}
In our experiments, the software we employed PyTorch Geometric 11.1 \cite{2019arXiv190302428F}, a key framework for deep learning on graphs, alongside PyTorch \cite{NEURIPS2019_bdbca288}, known for its flexibility and efficiency in building deep learning models. 
Additionally, PyGCL \cite{zhu2021empirical} played a crucial role, offering advanced tools for graph contrastive learning.
\par
For hardware, we leveraged a Nvidia GeForce RTX 3090 with 24 GB RAM, chosen for its advanced computational power, suitable for the demands of graph contrastive learning. 
Complementing this, the Intel(R) Xeon(R) CPU E5-2678 v3 @ 2.50GHz provided the necessary processing strength. 
Operating on Ubuntu 22.04 LTS, our setup was tailored for high efficiency and reliability, essential in managing extensive data and complex neural network operations. 
This configuration ensured a blend of speed and accuracy in our computational experiments.

\subsection{Hyper-parameter Selection}
\label{sec:c3:hyperpara}  
The selection of hyperparameters plays a pivotal role in determining the efficacy and efficiency of the learning process. Unlike model parameters, hyperparameters are not learned from the data but are set before the training phase and remain constant during the learning process.	
These parameters, which include the learning rate, the temperature parameter in the contrastive loss function, drop ratios, and the size of the hidden layers,  substantially impact the model's ability to learn meaningful representations.
\par
Table \ref{tab:HyeperParaNegAmplify} presents the hyper-parameter settings for the experiments performed in this chapter. 
$E_d$ refers to the edge drop ratio, and $F_m$ is the  feature masking ratio, where the superscript 1 or 2 refers to view one and view two. $\tau$ is the temperature parameter; training epochs, learning rates, weight decay, and size of hidden layers are also mentioned in the table. 


\begin{table}
	\centering
	\caption{Hype-parameter settings for NegAmplify. }
	\label{tab:HyeperParaNegAmplify}       
	\begin{tabular}{llllllllll}
		\hline\noalign{\smallskip}
		Dataset & $E_d^1$ & $F_m^1$ & $E_d^2$ & $F_m^2 $ & $\tau$ &\begin{tabular}{@{}c@{}}Training\\ epochs\end{tabular}& \begin{tabular}{@{}c@{}}Learning \\ rate\end{tabular}&\begin{tabular}{@{}c@{}}Weight \\ decay\end{tabular}&\begin{tabular}{@{}c@{}}Encoder \\ layers\end{tabular}\\
		\noalign{\smallskip}\hline\noalign{\smallskip}
		Cora&0.45&0.35&0.15& 0.5& 0.4&1200&$5\times10^{-3}$&$1^{-5}$&128,128\\
		CiteSeer&0.95&0.85&0.3& 0.25& 0.2&1200&$5\times10^{-4}$&$1^{-5}$&128,128\\	
		PubMed&0.5&0.45&0.4& 0.4&0.1&2000&$5\times10^{-4}$&$1^{-5}$&128,128\\
		DBLP&0.5&0.25&0.3& 0.45& 0.5&1200&$5\times10^{-4}$&$1^{-5}$&128,128\\
		\noalign{\smallskip}\hline\noalign{\smallskip}
		WikiCS &0.35 & 0.25&0.75 & 0.4& 0.85 &$1200$&$5\times10^{-4}$&$1^{-5}$&256,256\\               
		Am.Comp &0.5&0.4&0.15& 0.25& 0.15&$1200$&$5\times10^{-4}$&$1^{-5}$&256,256\\
		Am.Photo&0.1&0.15&0.45& 0.2& 0.5&$1200$&$1\times10^{-5}$&$1^{-5}$&256,256\\		
		Co.CS&0.2&0.5&0.5& 0.4& 0.7&$1200$&$1\times10^{-5}$&$1^{-5}$&256,256\\
		Actor&0.5&0.35&0.3& 0.5& 0.2&1200&$5\times10^{-4}$&$1^{-5}$&128,128\\
		\noalign{\smallskip}\hline
	\end{tabular}
\end{table}


\end{document}